\PassOptionsToPackage{table,xcdraw,dvipsnames}{xcolor}
\documentclass[10pt,twocolumn,letterpaper]{article}

\usepackage[pagenumbers]{cvpr}      

\newcommand{\red}[1]{{\color{red}#1}}

\usepackage{amsmath,amsfonts,bm}

\def\eqref#1{equation~\ref{#1}}

\def\1{\bm{1}}

\DeclareMathAlphabet{\mathsfit}{\encodingdefault}{\sfdefault}{m}{sl}
\SetMathAlphabet{\mathsfit}{bold}{\encodingdefault}{\sfdefault}{bx}{n}

\definecolor{cvprblue}{rgb}{0.21,0.49,0.74}
\usepackage[pagebackref,breaklinks,colorlinks,allcolors=cvprblue]{hyperref}

\usepackage{graphicx} 
\usepackage{url}
\newcommand{\indep}{\perp\!\!\!\perp}
\usepackage{multirow}

\newcommand\blue{\textcolor{blue}}
\newcommand\lightblue{\textcolor{NavyBlue}}
\definecolor{mygray}{gray}{.9}
\usepackage[capitalize]{cleveref}

\usepackage{amsmath,amstext,amsfonts,bm,amssymb,mathtools,amsthm}
\usepackage{tcolorbox}
\usepackage{color,xspace}
\usepackage{booktabs}
\usepackage{thm-restate}
\usepackage{bbding}
\usepackage{pifont}
\usepackage{wasysym}
\usepackage{wrapfig}

\usepackage[inline]{enumitem}
\theoremstyle{plain}
\newtheorem{theorem}{Theorem}[section]

\theoremstyle{definition}

\newtheorem{hypothesis}[theorem]{Hypothesis}
\theoremstyle{remark}

\usepackage{caption}

\usepackage{algorithm2e}

\def\paperID{7692} 
\def\confName{ICCV}
\def\confYear{2025}

\title{
Towards a Universal 3D Medical Multi-modality Generalization via Learning Personalized Invariant Representation
}

\author{
Zhaorui Tan$^{1,2,3}$, Xi Yang$^{2*}$, Tan Pan$^{1,4}$, Tianyi Liu$^{2,3}$, Chen Jiang$^{1,4}$, \\
Xin Guo$^{1,4}$, Qiufeng Wang$^{2}$, Anh Nguyen$^{3}$, Yuan Qi$^{1,4,6}$, Kaizhu Huang$^{5}$, Yuan Cheng$^{1,4}$\thanks{*Corresponding authors. This research was conducted during an internship at the Shanghai Academy of Artificial Intelligence for Science.}
\\
$^{1}$Shanghai Academy of Artificial Intelligence for Science, $^{2}$Xi'an Jiaotong-Liverpool University
\\$^{3}$University of Liverpool, $^{4}$AI$^3$ Fudan University,
$^{5}$Duke Kunshan University, $^{6}$Zhongshan Hospital
\\
{\tt\small  raytan@liverpool.ac.uk, Xi.Yang01@xjtlu.edu.cn, chengyuan@sais.com.cn}
}

\begin{document}
\maketitle

\begin{abstract}
Variations in medical imaging modalities and individual anatomical differences pose challenges to cross-modality generalization in multi-modal tasks. Existing methods often concentrate exclusively on common anatomical patterns, thereby neglecting individual differences and consequently limiting their generalization performance. 
This paper emphasizes the critical role of learning individual-level invariance,  i.e., personalized representation $\mathbb{X}_h$,  to enhance multi-modality generalization under both homogeneous and heterogeneous settings.
It reveals that mappings from individual biological profile to different medical modalities remain static across the population, which is implied in the personalization process.
We propose a two-stage approach: pre-training with invariant representation $\mathbb{X}_h$ for personalization, then fine-tuning for diverse downstream tasks.
We provide both theoretical and empirical evidence demonstrating the feasibility and advantages of personalization, showing that our approach yields greater generalizability and transferability across diverse multi-modal medical tasks compared to methods lacking personalization.  Extensive experiments further validate that our approach significantly enhances performance in various generalization scenarios.

\end{abstract}

\section{Introduction}


\begin{figure}
\centering
\includegraphics[width=0.8\linewidth]{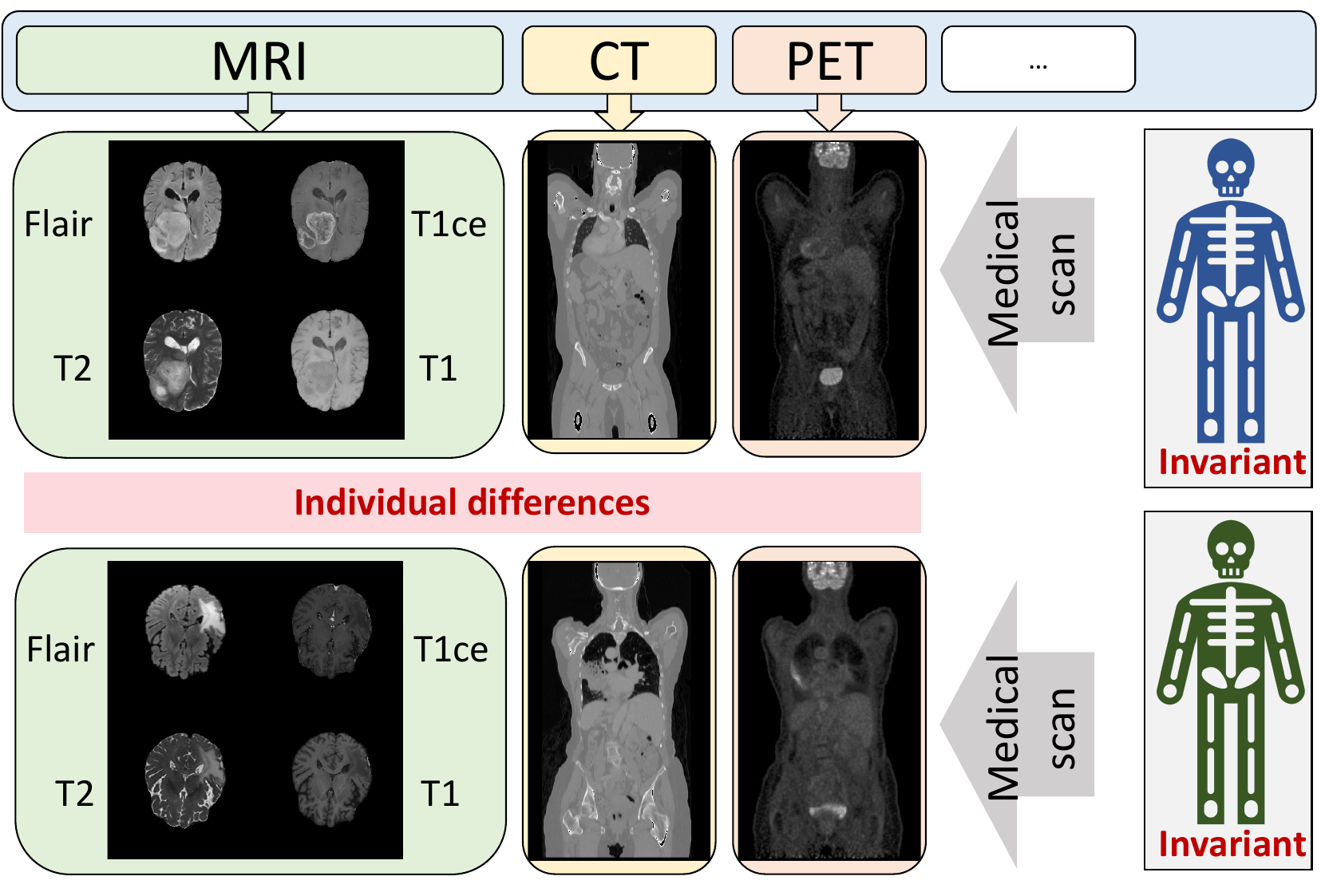}
\caption{A diagram of medical modalities and individual differences. These variations are significant and deserve further study by the medical intelligence community.}
\label{fig:banner}
\vspace{-0.4cm}
\end{figure}

Three-dimensional radiological medical images, generated through specialized techniques and radiopharmaceuticals, excel at highlighting specific physiological features, collectively providing a comprehensive view of a patient's structural and functional characteristics.
As illustrated in \cref{fig:banner}, current research in medical intelligence mainly targets structural modalities like Magnetic Resonance Imaging (MRI)~\citep{zhao2022modality} and Computed Tomography (CT)~\citep{ozbey2023unsupervised,zhan2024medm2g} scans. 
Other studies~\citep{yousefirizi2021toward} focus on the functional modalities associated with biochemistry, like Positron Emission Tomography (PET) scans. 
{Notably, all modalities exhibit significant individual variations.} Individuals differ fundamentally from population averages~\citep{whitcomb2012personalized}, making anatomical and metabolic variations evident.
In clinical settings, financial or physical limitations may restrict access to certain modalities, highlighting the need for generalization in medical image analysis across different modalities and individuals. The generalization faces significant challenges due to modality and individual variations. We classify generalization tasks into two types: \textit{homogeneous generalization} within structural or functional modalities, e.g., MRI sequences \cref{fig:banner}; and \textit{heterogeneous generalization} across both structural and functional modalities, such as CT and PET scans.

%

A well-generalized medical model should integrate insights from all available modalities to support downstream tasks, relying on condition invariance~\cite{cho2022cooperative,tan2024rethinking}.
Some methods enhance generalization through class-level anatomy invariance~\citep{jiang2023anatomical}, which may not fit models based on functional biochemistry and may degrade transferability for unseen individual domains. Single-modal task transfer research~\citep {tang2022self,wu2024voco,jiang2023anatomical} might not translate to multi-modal scenarios. Homogeneous generalization in medical tasks often employs cross-modality transfer~\citep{liu2023one,kim2024adaptive, zhan2024medm2g} or tackles missing modalities~\citep{liu2021incomplete,chen2023query,qiu2023scratch,qiu2023modal,zhan2024medm2g} in MRI or CT, while heterogeneous generalization is insufficiently explored.
Furthermore, these limited efforts often focus on modality or class-level invariance, neglecting individual differences and impeding a model's capacity to generalize effectively across patient populations and modalities.

\begin{figure}
\centering
\includegraphics[width=0.88\linewidth]{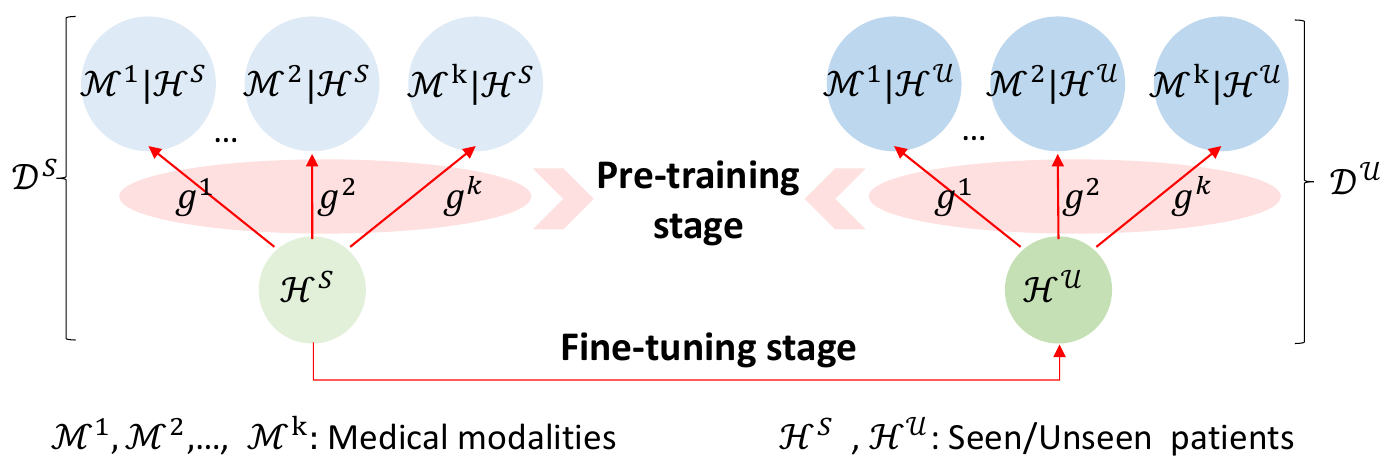}
\vspace{-0.3cm}
\caption{A diagram of patient domains and medical modalities linked to our method. In the Pre-training stage, static mappings between domains and modalities are implicitly learned. The fine-tuning stage addresses gaps between seen and unseen patients.
}
\label{fig:theory}
\vspace{-0.4cm}
\end{figure}

This paper reveals that both homogeneous and heterogeneous settings for multi-modality generalization can be tackled within the framework of personalization, i.e., learning the personal-level invariant representation. Based on medical imaging principles, mappings from an individual's identical biological profile to various modalities remain static across individuals. Learning personal-level invariant representations, which apply across multiple modalities, enhances cross-modality generalization and transferability to novel individuals. To formalize this approach, the hypothesis of a personalized invariant representation is introduced, denoted as $\mathbb{X}_h$, which exists for each individual in multi-modal generalization. Hypothesis ~\ref{hypo:X_h} rigorously details the constraints and properties of $\mathbb{X}_h$.

Building on this hypothesis, a general approach termed PUIR is proposed to augment the generalization of various medical imaging tasks through personalization. 
As shown in \cref{fig:theory}, our approach consists of two stages:
(1) The \textbf{pre-training stage} constructs an approximation of $\mathbb{X}_h$ using the learnable biological prior knowledge $\mathbb{O}$, via decomposition, invariance, and equivariance constraints during pre-training (refer to \cref{sec:constraints}), focusing on individual-level modality generalization.
(2) The \textbf{fine-tuning stage}, employs standard downstream training methods to adapt the modality encoders derived from the first stage to mitigate the gap between seen and unseen individuals.
A key distinction between our method and previous pre-training strategies lies in our specific focus on enhancing generalization across various medical modalities and individuals.

Importantly, this paper theoretically demonstrates that obtaining a personalized invariant representation, $\mathbb{X}_h$, is feasible through our approach, and such invariance leads to generalization improvements across various medical tasks (see~\cref{app:More theoretical analysis}).  
To validate our methodology, we conduct experiments on modality transfer and missing modality segmentation tasks, addressing not only the homogeneous generalization of MRI but also the rarely explored heterogeneous generalization, such as generalization between PET and CT. Our findings reveal that our approach successfully captures comprehensive, personalized information even when only partial modalities are available for a given individual.
Moreover, extensive experiments on both homogeneous (\cref{sec:homogeneous_modalities}) and heterogeneous (\cref{sec:heterogeneous_modalities,sec:special_case}) generalization demonstrate that our approach can be adapted for downstream tasks and surpasses current state-of-the-art (SOTA) methods in multiple tasks, validating its superiority. 
Our code and data splits are available at \url{https://github.com/zhaorui-tan/PUIR_ICCV25}.


\section{Related work}


{Medical generalization tasks} currently concentrates on 
homogeneous generalization, introducing tasks, such as modality transfer and missing modality segmentation for structural modalities — Flair, T1, T2, and T1ce of MRI- in brain tumor segmentation~\citep{zhao2022modality}, or between MRI and CT~\citep{zhan2024medm2g} for modality transfer. \cite{pan2023revealing} proposes an approach for heterogeneous generalization in terms of modality transfer, but only tailored for transferring PET to CT. 
In terms of tasks, most current studies focus on one specific generalization task, either segmentation~\citep{chen2021learning,wang2023prototype,ding2021rfnet,zhang2022mmformer,wang2021acn,liu2024mind11111} or modality transfer~\citep{isola2017image,zhu2017unpaired,fu2019geometry,park2020contrastive,kong2021breaking,liu2023one,shi2023m,dhariwal2021diffusion,ozbey2023unsupervised,kim2024adaptive,xing2024cross,zhan2024medm2g}.
This paper aims to develop an approach that is feasible for different downstream tasks under both homogeneous and heterogeneous modality generalization.
Our approach aims to learn the $\mathbb{X}_h$ through pre-training; we list related medical pre-training work 
\cite{tang2022self,wu2024voco,chen2020mocov2,jiang2023anatomical,pan2025structure} here. A notable work among them is \cite{jiang2023anatomical}, which extracts class-specific anatomical invariance. However, they only focus on a single modality. Such single-modality approaches may not be able to construct $\mathbb{X}_h$ for improving the generalization across modalities. 
The generalization in medical images also connects to alignment in multi-domain generalization for natural images~\cite{ganin2016domain,li2018deep,li2018domain,hu2020domain,tan2024rethinking}. 
Please refer to \cref{app:Related work} for a detailed literature review. 

\begin{figure*}
    \includegraphics[width=0.77\linewidth]{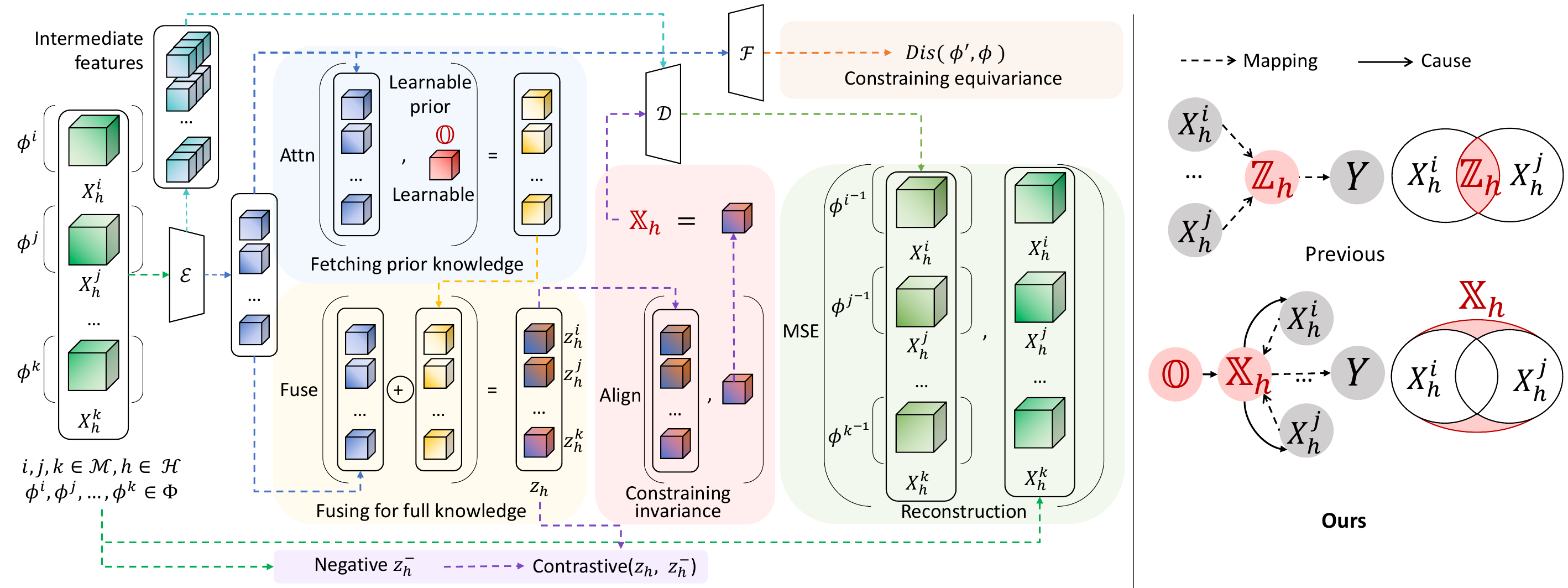}
    \centering
    \vspace{-0.3cm}
    \caption{Left: Overall framework of learning $\mathbb{X}_h$ during the pre-training stage.
    Right: Diagrams of differences between previous methods of learning $\mathbb{Z}_h$ and our proposed method of learning $\mathbb{X}_h$.}
    \label{fig:Framework}
    \vspace{-0.4cm}
\end{figure*}

\section{Learning personalized invariant representation for medical generalization}

\textbf{Notations.} 
The encoder is denoted to $\mathcal{E}$, while its associated decoder is defined as $\mathcal{D}$. An individual human entity, denoted as $h \in \mathcal{H}$, with corresponding medical images $X_h = {X^i_h, X^j_h, \dots, X^k_h}$, whereby $i, j, \dots, k \in \mathcal{M}$, and $\mathcal{M}$ cover all conceivable modality combinations. Intermediate features generated by $\mathcal{E}(X_h)$ and $\mathcal{E}(X^i_h)$ are designated as $x_h$ and $x^i_h$, respectively. The encoder's final layer features are $z_h$ and $z^i_h$. The learned approximation of $\mathbb{X}_h$ is expressed as ${\mathbb{X}_h}'$. The geometric warping function is $\phi^i \in \Phi$, with $\phi^i(X^i_h) \in \mathcal{X}$ and $\Phi$ as the ensemble of all possible warping functions. Conclusively, $I(\cdot; \cdot)$ and $P(\cdot)$ signify mutual information and probability distribution, respectively. 

\textbf{Personalized invariant representation hypothesis.}
Before addressing the problem for both homogeneous and heterogeneous generalization, we introduce the personalized invariant representation hypothesis, termed as  $\mathbb{X}_h$ Hypothesis, for medical imaging:
\begin{hypothesis}[$\mathbb{X}_h$ Hypothesis]
\label{hypo:X_h}
     Consider the set $\mathcal{M}$ of all possible modality combinations and the set $\Phi$ of possible geometric transformations (e.g., SO(3)), such as rotations corresponding to different postures of an individual. For a given individual from the population $h \in \mathcal{H}$, there exists a personalized invariant representation $\mathbb{X}_h$, which can be decomposed into modality-specific images 
    $X^i_h$ conditioned on a modality combination $i \in \mathcal{M}$. This relationship is formally expressed as: 
    \begin{equation}
    \label{eq:Grand_X_capability}
        X^i_h = \mathbb{X}_h|i; \; i\in \mathcal{M}, h\in \mathcal{H},\quad \textit{s.t.,} \mathbb{X}_h \indep \mathcal{M}, \Phi.
    \end{equation}
\end{hypothesis}
Despite potential differences in modalities and individual variations, clinical diagnoses focus on the biological conditions of a certain patient, which remain mostly invariant during a single hospital visit. 
Thus, the $\mathbb{X}_h$ Hypothesis holds in most cases. 
Our method aims to obtain an accurate approximation of $\mathbb{X}_h$. The overall learning framework for $\mathbb{X}_h$ is illustrated on the left-hand side of \cref{fig:Framework}. 
Data from each modality are encoded by $\mathcal{E}$, and the encoded features are used to retrieve knowledge from the learnable biological prior $\mathbb{O}$. The features and retrieved knowledge are then fused. By applying constraints of decomposition, equivariance, and invariance on the fused features, we approximate $\mathbb{X}_h$ effectively.


As illustrated in \cref{fig:Framework} right-hand side top, previous approaches~\citep{liu2021incomplete,chen2023query,qiu2023scratch,qiu2023modal} learn invariant representations $\mathbb{Z}_h$ across modalities through the encoder $\mathcal{E}$ for generalization: $\mathcal{E}(X^m_h) \to \mathbb{Z}_h, \mathbb{Z}_h \indep \mathcal{M}, m\in \mathcal{M}, h \in \mathcal{H}$ during pre-training or training.
Is $\mathbb{Z}_h$ a good approximation of $\mathbb{X}_h$, and does it benefit the generalization of different downstream tasks? The answer might be negative because 
such an approach may erase modal-specific information in $\mathbb{Z}$, making it impossible to be decomposed back into different modalities as shown in \cref{eq:Grand_X_capability}. 
Moreover, while current studies~\cite{havaei2016hemis,varsavsky2018pimms,zhang2022mmformer,ding2021rfnet} also disentangle modality-dependent features alongside the invariant representation $\mathbb{Z}$ to enhance transferability, this strategy may compromise the generalization ability of $\mathbb{Z}$. The reason is that the transferred targets become constrained by the learned modal-dependent features, potentially limiting their broader applicability.


\subsection{Pre-training stage}
\label{sec:learn_prior}

To learn a better approximation of $\mathbb{X}_h$,  
we leverage a learnable biological prior, denoted as $\mathbb{O}$.
If $\mathbb{O}$ can be learned, representations from any modality can complete themselves by retrieving the missing knowledge from $\mathbb{O}$, forming a better approximation of $\mathbb{X}_h$. 
Empirically, we initialize a learnable tensor as $\mathbb{O}$. As shown in \cref{fig:Framework}, the representation $z^i_h$ retrieves its missing knowledge from $\mathbb{O}$ via attention: ${z^i_h}' := attn(query:z^i_h, key:\mathbb{O}, value:\mathbb{O})$. The original representation and the retrieved knowledge are then fused through convolution: $\mathbb{X}^i_h:= conv ({z^i_h}', z^i_h)$. If the model is well-trained under the constraints of equivariance, invariance, and decomposition, the fused feature $\mathbb{X}^i_h$ becomes $\mathbb{X}'_h$, a good approximation of $\mathbb{X}_h$. The details of these constraints are discussed in \cref{sec:constraints}.





\subsubsection{Constraints of equivariance, invariance, and decomposition}
\label{sec:constraints}

\textbf{Contrastive learning.} Before we introduce the constraints, we include the contrastive loss as our baseline. During the pre-training stage, we follow previous work~\citep{chen2020mocov2,tang2022self} and employ the contrastive learning loss. 
Specifically, the positive pairs are constructed as augmented samples from the same sub-volume, while the negative pairs are the views from different sub-volumes.
Similar to~\citep{tang2022self}, the contrastive coding is obtained by attaching a linear layer $\psi(\cdot)$ to the $z_h$, its positive pair $z^+_h$, and all samples $\{z_h^i\}_{i=1}^{B}$ where $B$ is the total number of samples.
The contrastive loss is then defined as:
\begin{equation}
      \mathcal{L}_{contr}=-\log \frac{\exp \left(\operatorname{sim}\left( \psi(z_h), \psi(z^+_h)\right) / t\right)}{\sum_{i=0}^{B}\exp \left(\operatorname{sim}\left(\psi(z_h), \psi(z^i_h)\right) / t\right)},
\end{equation}
where $t$ is the measurement of the normalized temperature scale and $sim(\cdot, \cdot)$ denotes the dot product between normalized embeddings as the similarity.

As discussed in \cref{sec:learn_prior}, the $\mathbb{X}_h$ can be obtained through a model trained under the constraints of equivariance, invariance, and decomposition. 
The following part presents details of those constraints according to the $\mathbb{X}_h$ hypothesis. 

\textbf{{Invariance constraint.}} 
We constrain 
the invariance for $\mathbb{X}_h$ where $\mathbb{X}_h \indep \mathcal{M},\Phi$ through alignment. 
The $ z^i_h $ firstly uses attention to fetch the knowledge from the prior: ${z^i_h}' = attn(z^i_h,\mathbb{O})$ and then they are concatenated and fused through convolution ${\mathbb{X}^i_h}' = conv(z^i_h \oplus  {z^i_h}' )$.
Despite the different modality combinations and geometric transformations, $\mathbb{X}_h$ should be invariant for the person:
\begin{equation}
    \mathcal{L}_{inv} = \sum\nolimits  \left \| {\mathbb{X}^i_h}' , {\mathbb{X}_h}'  \right \| ^2, \quad i \in \mathcal{M}.
\end{equation}
While it is well aligned, ${\mathbb{X}^i_h}' = {\mathbb{X}^j_h}'= ... ={\mathbb{X}_h}'$ where $j \in \mathcal{M}$. Empirically, we use ${\mathbb{X}_h}' \triangleq mean({\mathbb{X}^i_h}', {\mathbb{X}^j_h}', ...)$ and $mean(\cdot)$ refers the averaging of the input sequence.

\textbf{{Equivariance constraint.}} 
To learn better $\mathbb{O}$ and ${\mathbb{X}_h}'$  as the personalized invariant representation, we constrain the geometric equivariance and representation invariance.
Consider the sample space of all modalities $X^i_h \in \mathcal{X}, i \in \mathcal{M}$,  
the geometric equivariance constraint forces that the geometry of the generated medical image is equivariant to $\phi^i$, which can be constrained by the MSE loss in \cref{eq:decom_loss}.
Furthermore, such equivariance demands that $\phi(x^i_h)$ and $z^i_h$ contain the information of the geometric transformation $\phi^i$, inferring that it is able to extract the $\phi^i$ from $\phi^i(x^i_h)$ and $z^i_h$. 
Therefre, if $\phi^i$ can be extracted from the last-layer output $z^i_h$, it can also be extracted from the $\phi^i(x^i_h)$ from the previous layers:
\begin{equation}
    \min\nolimits_{\mathcal{D},\mathcal{F}} Dis( \phi^i,\mathcal{F}(z^i_h)),
\end{equation}
where $\mathcal{F}:\mathcal{F}(z^i_h) \to {\phi^i}'$ extracts the geometric transformation and $Dis(\dot,\dot)$ denotes the distance measurement between ${\phi^i}'$ and $\phi^i$. 
Empirically, following~\citep{tang2022self}, we also adopt rotation as the geometric transformation, predicting the angle categories of input sub-volume is rotated. Under this case, $\Phi_R$ is defined as rotations at $[0, 90, 180, 270]$ degrees along the z-axis, and $ \phi_r^i \in \Phi_R$ is the ground truth rotation categories. 
$\mathcal{F}{z^i_h}$ produces the softmax probabilities of rotation categories, predicting which kind of rotation is applied, and loss is in the form of:
\begin{equation}
    \mathcal{L}_{equ} = - \sum\nolimits_{r=1}^{|\Phi_R|} \phi_r^i \log \mathcal{F}(z^i_h).
\end{equation}

\textbf{Decomposition constraint.}
As shown in \cref{eq:Grand_X_capability} of $\mathbb{X}_h$ Hypothesis, the ${\mathbb{X}_h}'$ need to be able to be decomposed as different modalities, which refers: $ \min_{\mathcal{E},\mathcal{D},\mathbb{O}}I(P({\mathbb{X}_h}'|i); P(X^i_h))$. 
An intuitive approach is reconstructing all possible modalities by using $\mathbb{X}_h$, whose objective can be formed as:
\begin{equation}
\label{eq:decom_loss}
    \mathcal{L}_{decom} \!=\! \sum\nolimits_{1}^{|\mathcal{M}|} \left \|
    {\phi^{i}}^{-1} \! \left (\mathcal{D} \left ({\mathbb{X}_h}'| \phi^i(x^i_h) \right ) \right ), X_h
    \right \|^2 , 
\end{equation}
where $i\in \mathcal{M}$ and $\phi^i(x^i_h)$ represents intermediate representations produced during $\mathcal{E}(\phi^i( X^i_h))$ and $X_h$ denotes all possible modalities. Intuitively, $\phi^i(x^i_h)$ from earlier layers of the encoder constrains modality information thus $\mathcal{D}  ({\mathbb{X}_h}'| \phi^i(x^i_h)) \triangleq \mathcal{D}  ({\mathbb{X}_h}'|i) $. 
Specifically, the generated medical image is transformed back by using the inverse of $\phi^i$ to align with the inputs.

\textbf{Final loss for learning $\mathbb{X}_h$.} The final loss for pre-training is the combination of the above losses:
\begin{equation}
    \mathcal{L}_{pre} = \mathcal{L}_{contr} + \mathcal{L}_{decom} + \mathcal{L}_{equ} + \mathcal{L}_{inv},
\end{equation}
where the weight of each loss is omitted here. 


\begin{table*}[!t]
\centering
\begin{minipage}{0.33\textwidth}
\centering
\resizebox{\linewidth}{!}{%
\begin{tabular}{l|ccc}
\hline
\multicolumn{1}{l|}{\textbf{Task}} & \multicolumn{3}{c}{\textbf{T1→T2}} \\ \hline
Method & {PSNR↑} & {NMSE↓}  & {SSIM↑}  \\ \hline
&\multicolumn{3}{c}{2D} \\ \hline
Pix2Pix~\citep{isola2017image}  & 24.624$^*$& 0.109$^*$& 0.874$^*$\\
CycleGAN~\citep{zhu2017unpaired} & 23.535$^*$& 0.155$^*$ & 0.837$^*$\\
NICEGAN~\citep{chen2020reusing} & 23.721$^*$ & 0.148$^*$ & 0.840$^*$ \\
RegGAN~\citep{kong2021breaking} & 24.884$^*$ & 0.094$^*$ & 0.881$^*$  \\
ResViT~\citep{dalmaz2022resvit} & 25.578$^*$& 0.088$^*$& 0.895$^*$ \\ 
\hline
&\multicolumn{3}{c}{3D} \\ \hline
Pix2Pix & 23.740$^*$ & 0.138$^*$ & 0.835$^*$ \\
CycleGAN & 25.181$^*$& 0.097$^*$& 0.887$^*$\\
EaGAN~\citep{yu2019ea} & 24.884$^*$ & 0.094$^*$ & 0.881$^*$  \\
 \cellcolor{mygray}\textbf{Ours (PUIR)} & \cellcolor{mygray}\blue{\textbf{30.756}} & \cellcolor{mygray}\blue{\textbf{0.065}} & 
\cellcolor{mygray}\blue{\textbf{0.944}}   \\ \hline
\end{tabular}%
}
\end{minipage}%
\hfill
\begin{minipage}{0.29\textwidth}
\centering
\resizebox{\linewidth}{!}{%
\begin{tabular}{l|ccc}
\hline
\multicolumn{1}{l|}{\textbf{Task}} & 
\multicolumn{3}{c}{\textbf{T2 → Flair}} \\ \hline
Method  & \multicolumn{1}{c}{PSNR↑} & \multicolumn{1}{c}{NMSE↓} & {SSIM↑} \\ \hline
&\multicolumn{3}{c}{2D} \\ \hline
Pix2Pix~\citep{isola2017image} & 24.82$^\dag$ &  0.0250$^\dag$ &  0.846$^\dag$  \\
CycleGAN~\citep{zhu2017unpaired} & 23.418$^*$& 0.164$^*$ & 0.825$^*$  \\
NICEGAN~\citep{chen2020reusing} & 23.643$^*$ & 0.148$^*$ & 0.829$^*$ \\
RegGAN~\citep{kong2021breaking} & 24.576$^*$ & 0.112$^*$ & 0.852$^*$  \\
ResViT~\citep{dalmaz2022resvit} & 24.825$^*$ & 0.108$^*$& 0.861$^*$ \\ 
Diffusion~\citep{dhariwal2021diffusion} & 31.98$^\dag$ & -& 0.930$^\dag$  \\
MD-Diff~\citep{xing2024cross}& 30.76$^\dag$ & - & 0.934$^\dag$ \\
\hline
&\multicolumn{3}{c}{3D} \\ \hline
Pix2Pix & 23.508$^*$ & 0.152$^*$ & 0.822$^*$ \\
CycleGAN& 24.602$^*$& 0.113$^*$ & 0.854$^*$  \\
EaGAN~\citep{yu2019ea}  & 24.576$^*$ & 0.112$^*$ & 0.852$^*$ \\
\cellcolor{mygray}\textbf{Ours (PUIR)} & \cellcolor{mygray}\blue{\textbf{32.224}} & \cellcolor{mygray}\blue{\textbf{0.046}} & \cellcolor{mygray}\blue{\textbf{0.941}}  \\ \hline
\end{tabular}%
}
\end{minipage}%
\hfill
\begin{minipage}{0.26\textwidth}
\centering
\resizebox{\linewidth}{!}{%
\begin{tabular}{l|cc}
\hline
\multicolumn{1}{l|}{\textbf{Task}}  & \multicolumn{2}{c}{\textbf{T1 → T1ce}}\\ \hline
Method & \multicolumn{1}{c}{PSNR↑} & \multicolumn{1}{c}{SSIM↑} \\ \hline
&\multicolumn{2}{c}{2D} \\ \hline
Pix2Pix~\citep{isola2017image}   & 27.05$^\dag$  & 0.858$^\dag$ \\
CycleGAN~\citep{zhu2017unpaired} & 30.13$^\dag$  & 0.906$^\dag$ \\
GcGAN~\citep{fu2019geometry} & 25.98$^\dag$ &0.872$^\dag$  \\
CUT~\citep{park2020contrastive}  &  26.27$^\dag$ & 0.846$^\dag$ \\
RegGAN~\citep{kong2021breaking} & 31.36$^\dag$ & 0.930$^\dag$  \\
ResViT~\citep{dalmaz2022resvit} & 31.46$^\dag$ &0.932$^\dag$\\ 
Diffusion~\citep{dhariwal2021diffusion}  & 29.22$^\dag$ & 0.921$^\dag$   \\
MD-Diff~\citep{xing2024cross} & 33.08$^\dag$  & 0.948$^\dag$  \\
\hline
&\multicolumn{2}{c}{3D} \\ 
\hline
 \cellcolor{mygray}\textbf{Ours (PUIR)} & \cellcolor{mygray}\blue{\textbf{34.547}} & \cellcolor{mygray}\blue{\textbf{0.955}}  \\ \hline
\end{tabular}%
}
\end{minipage}
\vspace{-0.2cm}
\caption{\textbf{Modality transfer results of MRI} on BRATS23: Comparison between previous \textbf{one-to-one modality transfer methods} and our method. 
The best results are highlighted in \blue{blue}. 
Results denoted with $^*$ are gained from~\cite{kim2024adaptive}; with $^\dag$ are gathered from \cite{xing2024cross}.
}
\label{tab:gen_BRATS23_highlight}
\vspace{-0.4cm}
\end{table*}

\subsubsection{The connection between the constraints and global biological prior}
\label{sec:contraints}
It is important to note that the above constraints are closely interconnected, as they align with \cref{eq:Grand_X_capability}. After obtaining additional knowledge from $\mathbb{O}$, the invariance constraint ensures that the representations from each modality for a given individual are the same, such that ${\mathbb{X}^i_h}'$ and ${\mathbb{X}_h}'$ can be considered equivalent. Combined with the decomposition constraint, which enforces that ${\mathbb{X}_h}'$ is shared for the generation of all possible modalities, ${\mathbb{X}_h}'$ is thus able to generalize across modalities.
Additionally, the equivariance and decomposition constraints implicitly maintain SO(3)-equivariance by satisfying the relation $\mathcal{D} \circ \mathcal{E}(\phi^i(X^i_h); \theta) = \phi^i(\mathcal{D} \circ \mathcal{E}(X^i_h; \theta))$\footnote{The SO(3) transformations are left-multiplication; they are expressed here in a simplified form, using $\phi^i(\cdot)$.}, where $\theta$ represents the model parameters after training. 
This ensures that geometric transformations are preserved in the latent features $z^i_h$, such that $\mathcal{F}(z^i_h) = \phi^i$. The invariance constraint then requires that $z^i_h$ with geometric transformations can retrieve features from $\mathbb{O}$ to form ${\mathbb{X}^i_h}'$, which remains invariant to any geometric transformation.
This implicitly constrains $\mathbb{O}$ to contain comprehensive biological information, including other potential geometric transformations, thereby improving the $\mathbb{X}_h'$ through $\mathbb{O}$ in approximating $\mathbb{X}_h$ and enhancing the robustness of ${\mathbb{X}_h}'$.


\subsection{Fine-tuning stage}

After pre-training with the loss function $\mathcal{L}_{pre}$, the model is then utilized for downstream tasks such as segmentation or generation. We denote the commonly used loss functions for these tasks, such as dice loss, cross-entropy loss, or mean squared error (MSE) loss, as $\mathcal{L}_{ori}$, where paired data and labels $(X, Y) \in (\mathcal{X}, \mathcal{Y})$ are provided. In addition to $\mathcal{L}_{ori}$, we incorporate the invariance loss, denoted as $\mathcal{L}{inv}$, as part of the fine-tuning process for downstream tasks:
\begin{equation}
    \mathcal{L}_{down} = \mathcal{L}_{ori} + \mathcal{L}_{inv}.
\end{equation}


Empirically, we adopt the SwinUNETR architecture~\citep{hatamizadeh2022swin} as the backbone of the encoder $\mathcal{E}$, and implement the proposed components. The model is trained with $\mathcal{L}_{pre}$ during the pre-training phase; users have the option to either use the standard SwinUNETR by loading only our pre-trained encoder weights or to employ our proposed model structure with all pre-trained weights for downstream tasks. 
Notably, all modalities for a given individual, $X_h = {X^i_h, X^j_h, ..., X^k_h}, i,j,...,k \in \mathcal{M}$, share the same encoder, with the encoder's channel size set to match the number of modality types. The input volume size for all experiments is fixed at $96 \times 96 \times 96$.
Further empirical details on how $\mathbb{X}_h$ is leveraged for homogeneous and heterogeneous generalization are provided in \cref{sec:homogeneous_modalities,sec:heterogeneous_modalities}.

\subsection{Theoretical analysis}
\label{app:More theoretical analysis}


Consider a mapping  $ g: \mathcal{H} \to \mathcal{M} $ that extracts modalities from the individual biological profile, where $ g \in \mathcal{G} $ represents a set of candidate hypotheses.  
In modality generalization, our objective is to guarantee that the model effectively generalizes across diverse modalities $ \mathcal{M} $ for previously unseen patients. To achieve this, the learning bound must account for the relationships between seen $ \mathcal{H}^\mathcal{S} $ and unseen patients $ \mathcal{H}^\mathcal{U} $. {Inspired by the theoretical framework of
domain adaptation (DA) risk bounds 
in~\cite{ben2010theory}, we derive the following bound for medical cross-modality generalization:}  
\begin{theorem}
Let $ R^\mathcal{S} $ and $ R^\mathcal{U} $ denote the generalization errors on the modalities from the seen patient domain $ \mathcal{D}^\mathcal{S} $ and the unseen patient domain $ \mathcal{D}^\mathcal{U} $, respectively.  For a given hypothesis $ g \in \mathcal{G} $,  the overall unseen risk is bounded by:
\begin{equation}
\begin{split}
R^\mathcal{U}(g) \leq R^\mathcal{S}(g) + d_{\mathcal{G}\Delta\mathcal{G}}(\mathcal{D}^\mathcal{S},\mathcal{D}^\mathcal{U}) + \lambda ,
\end{split}
\end{equation}
where 
$d_{\mathcal{G}\Delta\mathcal{G}}$ is the $\mathcal{G}\Delta\mathcal{G}$-divergence between $\mathcal{D}^\mathcal{S}$ and $\mathcal{D}^\mathcal{U}$, 
and joint hypothesis $\lambda$  quantifies the inherent difficulty of aligning the seen and unseen domains (can be omitted).
\end{theorem}

The bound emphasizes the importance of minimizing the divergence between domains and achieving a low risk on the seen domain to ensure effective generalization to unseen domains. For a well-trained model on $R^\mathcal{S}(g)$ is minimized. the key focus shifts to $d_{\mathcal{G}\Delta\mathcal{G}}(\mathcal{D}^\mathcal{S},\mathcal{D}^\mathcal{U})$. Considering modalities,  $d_{\mathcal{G}\Delta\mathcal{G}}(\mathcal{D}^\mathcal{S},\mathcal{D}^\mathcal{U})$  is defined as follows:
\begin{equation}
\begin{split}
\nonumber
        d_{\mathcal{G}\Delta\mathcal{G}}(\mathcal{D}^\mathcal{S},\mathcal{D}^\mathcal{U}) \!:= d_{\mathcal{G}\Delta\mathcal{G}}([\mathcal{H}^\mathcal{S} \oplus  \mathcal{M}|\mathcal{H}^\mathcal{S}], [\mathcal{H}^\mathcal{U} \oplus  \mathcal{M}|\mathcal{H}^\mathcal{U}] ),
\end{split}
\end{equation}
{where $\oplus$ denotes that each domain combines the patient domain with its derived modalities. }
Upon rearranging, it follows that:
\begin{equation}
\label{eq:key_obj}
    d_{\mathcal{G}\Delta\mathcal{G}}(\mathcal{H}^\mathcal{S},\mathcal{H}^\mathcal{U}) + d_{\mathcal{G}\Delta\mathcal{G}}(\mathcal{M} | \mathcal{H}^\mathcal{S}, \mathcal{M} | \mathcal{H}^\mathcal{U}).
\end{equation}

\textbf{Our approach}  minimizes $ d_{\mathcal{G}\Delta\mathcal{G}}(\mathcal{M} | \mathcal{H}^\mathcal{S}, \mathcal{M} | \mathcal{H}^\mathcal{U}) $ by initially learning $ \mathbb{X}_{h} $ for each patient at the pre-training stage and then employing a fine-tuning stage to minimize $ d_{\mathcal{G}\Delta\mathcal{G}}(\mathcal{H}^\mathcal{S},\mathcal{H}^\mathcal{U}) $. 
As shown in \cref{fig:theory}, while each modality uniquely visualizes a patient's biological profile, the underlying generation principles remain unchanged. Thus, the mapping $ g_{m}: \mathcal{H} \to \mathcal{X}_{m} $ that derives the modality-specific visualizations $ m \in \mathcal{M}$ from the individual biological profile is consistent across individuals.  
Since direct access to a person’s anatomical structure is infeasible, we propose learning an invariant representation $ \mathbb{X}_h $ for $h \in \mathcal{H}$ with a biological prior, allowing decomposition into various modalities. To ensure $ \mathbb{X}_h $ effectively captures anatomical features, we enforce equivariance constraints to preserve geometric information. This learned representation augments the quality of $ g_{m} $, improving its generalization ability by reducing $ d_{\mathcal{G}\Delta\mathcal{G}}(\mathcal{M} | \mathcal{H}^\mathcal{S}, \mathcal{M} | \mathcal{H}^\mathcal{U})$.  
In the second fine-tuning stage, it is only necessary to mitigate $ d_{\mathcal{G}\Delta\mathcal{G}}(\mathcal{H}^\mathcal{S},\mathcal{H}^\mathcal{U}) $, thus facilitating the transfer process.

\textbf{Previous approaches} ignore personal variations, opting instead to focus on minimizing the gap between modalities, as noted in $ d_{\mathcal{G}\Delta\mathcal{G}}(\mathcal{M}^{\mathcal{S}}, \mathcal{M}^{\mathcal{U}})$. 
To facilitate a more effective comparison with our approach, we simplify this issue by positing that $d_{\mathcal{G}\Delta\mathcal{G}}$ is quantifiable via Kullback-Leibler (KL) divergence. 
Consequently, previous method aimed at minimizing $ d_{\mathcal{G}\Delta\mathcal{G}}(\mathcal{M}^{\mathcal{S}}, \mathcal{M}^{\mathcal{U}})$ can be converted as follows:
\begin{equation}
\begin{split}
\nonumber
        &d_{\mathcal{G}\Delta\mathcal{G}}(\mathcal{M}^{\mathcal{S}}, \mathcal{M}^{\mathcal{U}} ):=
  KL(\mathcal{M}^{\mathcal{S}} \Vert \mathcal{M}^{\mathcal{U}}) \\
    = &KL(\mathcal{H}^\mathcal{S} \Vert \mathcal{H}^\mathcal{U} )+ E_{\mathcal{H}^\mathcal{S}, \mathcal{H}^\mathcal{U} }[KL(\mathcal{M}^{\mathcal{S}}|\mathcal{H}^\mathcal{S} 
    \Vert \mathcal{M}^{\mathcal{U}}|\mathcal{H}^\mathcal{U})] \\
    \ge & KL(\mathcal{M}^{\mathcal{S}}|\mathcal{H}^\mathcal{S} 
    \Vert \mathcal{M}^{\mathcal{U}}|\mathcal{H}^\mathcal{U}) =: d_{\mathcal{G}\Delta\mathcal{G}}(\mathcal{H}^\mathcal{S},\mathcal{H}^\mathcal{U}) \; (\textbf{Ours}).
\end{split}
\end{equation}
Our methodology narrows the gap identified in \cref{eq:key_obj} compared to earlier methods, thereby demonstrating enhanced generalization and transferability capabilities.

\section{Experiments}

{\textbf{Overall experiment arrangement.} Our experiments examine various generalization scenarios in two stages: generation pre-training and downstream fine-tuning. We assess pre-training by comparing generation quality to SOTA medical generation methods and evaluate fine-tuning by comparing with SOTA methods for each task, including benchmarking against SOTA medical self-supervised pre-training methods. More results and analysis are in \cref{app:more_res}.}

\subsection{Homogeneous generalization: MRI}
\label{sec:homogeneous_modalities}

\begin{table}[t]
\centering
\resizebox{0.8\linewidth}{!}{%
            \begin{tabular}{c|lll}
            \hline
            From scratch& \multicolumn{1}{c}{PSNR↑} & \multicolumn{1}{c}{NMSE↓} & \multicolumn{1}{c}{SSIM↑} \\ \hline
            {SwinUNETR} & 29.1776 & 0.2196 & 0.8961 \\
            {MS-SPADE} & 26.4225 & 0.0822 & 0.9086 \\ \hline
            
            \rowcolor{mygray}\textbf{Ours (PUIR) ($96 \times 96 \times 96$)} & \blue{\textbf{35.4378}} & \blue{\textbf{0.0362}} & \blue{\textbf{0.9587}} \\
            \rowcolor{mygray}\textbf{Ours (PUIR) ($192\times 192 \times 192 $)} & 33.7418 & 0.0481 & 0.9502 \\ \hline
            \end{tabular}%
            }
            \vspace{-0.2cm}
\caption{\textbf{Modality transfer results of MRI} on BRATS23: 
Comparison between the previous \textbf{one-to-all modality transfer methods} and ours for transfer between all four modalities. 
The averaged results of metrics for all validation samples across all modalities are listed.
The best results are highlighted in \blue{blue}. 
Please refer to Appendix~\cref{tab:app_gen_BRATS23_main} for more detailed results.
}
\label{tab:gen_BRATS23_main}
\vspace{-0.4cm}
\end{table}

\subsubsection{Pre-training stage: Modality transfer}
\label{sec:seq_mod_trans}



\textbf{Experimental settings.} 
Following previous methods~\citep{kim2024adaptive}, we utilize the multi-modal brain tumor segmentation challenge 2023 (BRATS23) dataset~\citep{baid2021rsna,menze2014multimodal,bakas2017advancing,bakas2017segmentation}. BRATS23 includes four structural MRI modalities (T1, T1ce, T2, and FLAIR) for each individual. Our model is tested on the BRATS23 validation set, which contains these four modalities for 219 individuals.
We evaluate the quality of synthesis using peak signal-to-noise ratio (PSNR), normalized mean squared error (NMSE), and structural similarity index (SSIM)~\citep{yi2019generative}. 
To provide comprehensive results, we separately compare the translation results for T1 $\to$ T2 and T2 $\to$ FLAIR, as some previous methods are only capable of single-modality transfer.
These include both 2D and 3D generation methods, as shown in \cref{tab:gen_BRATS23_highlight}. 
Additionally, we employed SwinUNTER for multi-modality translation comparisons. All evaluations were performed on 3D volumes; for the 2D methods, synthesized target images were stacked to form a 3D volume for comparison. Please refer to model training details to \cref{app:exp}.

\begin{table}[t]
\centering
\centering
\resizebox{\linewidth}{!}{%
\begin{tabular}{c|cc|c|cc|cc|cc}
\hline
 & \multicolumn{2}{c|}{All settings} & FM & \multicolumn{2}{c|}{MN=1} & \multicolumn{2}{c|}{MN=2} & \multicolumn{2}{c}{MN=3} \\
 \multirow{-2}{*}{Method} & Mean & Std. & - & Mean & Std. & Mean & Std. & Mean & Std. \\ \hline
&\multicolumn{9}{c}{Tumor core}\\ \hline
 RFNET~\citep{ding2021rfnet} & 76.08 & 6.99 & 83.40 & 80.63 & 4.53 & 76.57 & 7.15 & 68.95 & 6.07 \\
 mmFormer~\cite{zhang2022mmformer} & 76.43 & 5.83 & 82.22 & 79.78 & 4.33 & 76.55 & 6.03 & 71.45 & 6.80 \\
 SPA~\citep{wang2023multi} & 74.80 & 6.95 & 82.23 & 78.99 & 5.38 & 75.01 & 8.31 & 68.44 & 8.88 \\
 M3AE~\citep{liu2023m3ae} & 72.67 & 7.43 & 80.29 & 77.61 & 4.59 & 73.37 & 7.96 & 64.79 & 9.85 \\
 M2F~\citep{shi2023m} & 73.69 & 6.83 & 80.34 & 77.48 & 5.19 & 74.17 & 6.88 & 67.51 & 6.63 \\
 \hline
\rowcolor{mygray} SwinUNETR pre-training & 79.19	& {{4.24}} &	84.69	&82.59	&3.43	&79.19	&2.69&	74.41	&{{3.54}}	\\
\cellcolor{mygray}\textbf{Ours (PUIR)} & \cellcolor{mygray}\blue{\textbf{79.78}} & \cellcolor{mygray}{{4.55}} & \cellcolor{mygray}\blue{\textbf{86.72}} & \cellcolor{mygray}\blue{\textbf{83.64}} & \cellcolor{mygray}{{2.29}} & \cellcolor{mygray}\blue{\textbf{79.56}} & \cellcolor{mygray}{{3.28}} & \cellcolor{mygray}\blue{\textbf{74.51}} & \cellcolor{mygray}{{3.87}} \\ \hline
&\multicolumn{9}{c}{Enhancing tumor}\\ \hline
RFNET & 59.31 & 15.10 & 73.65 & \blue{\textbf{66.91}} & 12.90 & 59.17 & 14.50 & 48.35 & 13.86 \\
 mmFormer & 62.14 & 18.60 & \blue{\textbf{79.91}} & 71.54 & 15.94 & 61.77 & 19.65 & 48.86 & 20.73 \\
SPA & 58.92 & 17.68 & 73.40 & 68.44 & 17.11 & 58.05 & 19.77 & 47.10 & 19.99 \\
M3AE & 55.98 & 17.45 & 73.79 & 65.09 & 14.54 & 55.53 & 20.37 & 43.09 & 22.03 \\
 M2F & 58.84 & 16.58 & 75.26 & 66.67 & 13.83 & 58.99 & 16.19 & 46.70 & 15.97 \\
  \hline
\rowcolor{mygray} SwinUNETR pre-training & 59.47	&{{6.25}}&	61.38&	60.61	&{{3.43}}&	60.85&	6.10	&55.79	&9.14\\
\cellcolor{mygray}\textbf{Ours (PUIR)} & \cellcolor{mygray}\blue{\textbf{63.49}} & \cellcolor{mygray}{{7.58}} & \cellcolor{mygray}70.64 & \cellcolor{mygray}64.44 & \cellcolor{mygray}{{6.62}} & \cellcolor{mygray}\blue{\textbf{63.87}} & \cellcolor{mygray}{{11.45}} & \cellcolor{mygray}\blue{\textbf{60.19}} & \cellcolor{mygray}{{11.31}} \\ \hline
&\multicolumn{9}{c}{ Whole tumor}
\\ \hline
RFNET & 83.92 & 6.14 & 89.27 & 87.25 & 2.94 & 84.95 & 3.81 & 77.70 & 7.46 \\
mmFormer & 84.84 & 5.35 & 88.26 & 87.59 & 2.40 & 85.36 & 5.60 & 80.45 & 4.64 \\
 SPA & 84.52 & 5.48 & 89.03 & 87.81 & {{1.25}} & 85.26 & 4.69 & 78.98 & 7.32 \\
M3AE & 81.52 & 6.71 & 86.82 & 85.64 & 1.36 & 82.43 & 6.08 & 74.74 & 8.69 \\
 M2F & 83.88 & 5.79 & 88.72 & 87.30 & 1.99 & 84.62 & 2.57 & 78.13 & 7.81 \\
  \hline
\rowcolor{mygray} SwinUNETR pre-training & 86.55	&3.49&	89.22&	88.18	&1.38&	86.69&	3.15&	84.04	&4.91\\
\cellcolor{mygray}\textbf{Ours (PUIR)} & \cellcolor{mygray}\blue{\textbf{87.63}} & \cellcolor{mygray}{{3.25}} & \cellcolor{mygray}\blue{\textbf{91.19}} & \cellcolor{mygray}\blue{\textbf{89.49}} & \cellcolor{mygray}1.82 & \cellcolor{mygray}\blue{\textbf{87.45}} & \cellcolor{mygray}{{1.03}} & \cellcolor{mygray}\blue{\textbf{85.17}} & \cellcolor{mygray}{{3.93}} \\ \hline
\end{tabular}%
}
\caption{\textbf{Missing modality segmentation results of MRI} on BRATS18: Comparisons between previous SOTA methods and ours. MN: Number of how many missing modalities; FM: Full Modality.  We report mean and standard deviations of DICE results under the MN. 
The best DICE results are highlighted in \blue{blue}.
Please refer to Appendix~\cref{tab:app_missing_seg_brats18_main} for more details. 
}
\label{tab:missing_seg_brats18_main}
\vspace{-0.5cm}
\end{table}


\begin{figure*}[t]
\centering
\begin{minipage}{0.79\linewidth}
\centering
\resizebox{\linewidth}{!}{%
\begin{tabular}{l|c|cc|ccc|c|cc|ccc}
\hline
 \multicolumn{1}{c}{} & \multicolumn{6}{c|}{SSIM↑} & \multicolumn{6}{c}{PSNR↑} \\ \hline
\begin{tabular}[c]{@{}l@{}}+ Contrastive \\+ Decomposition \end{tabular}& $\bullet $ & $\bullet $ & $\bullet $ & $\bullet $ & $\bullet $ & \cellcolor{mygray}$\bullet $ & $\bullet $ & $\bullet $ & $\bullet $ & $\bullet $ & $\bullet $ & \cellcolor{mygray}$\bullet $ \\ \hline
 + Equivariance  & $\bullet $ &  & $\bullet $ & $\bullet $ &  & \cellcolor{mygray}$\bullet $ & $\bullet $ &  & $\bullet $ & $\bullet $ &  & \cellcolor{mygray}$\bullet $ \\
+ Invariance  &  & $\bullet $ & $\bullet $ &  & $\bullet $ & \cellcolor{mygray}$\bullet $ &  & $\bullet $ & $\bullet $ &  & $\bullet $ & \cellcolor{mygray}$\bullet $ \\
+ $\mathbb{O}$ &  &  &  & $\bullet $ & $\bullet $ & \cellcolor{mygray}$\bullet $ &  &  &  & $\bullet $ & $\bullet $ & \cellcolor{mygray}$\bullet $ \\ 
 \hline
PET $\to$ PET & 0.9903 & 0.9835 & \lightblue{\textbf{0.9955}} & 0.9931 & {{0.9957}} & \cellcolor{mygray}\blue{\textbf{0.9969}} & 44.8811 & 42.2223 & {46.5603} & 45.5829 & \lightblue{\textbf{47.4198}} & \cellcolor{mygray}\blue{\textbf{49.5473}} \\
CT $\to$ CT &\lightblue{\textbf{0.9739}} & 0.9475 & 0.9419 & 0.9437 & 0.9664 & \cellcolor{mygray}\blue{\textbf{0.9780}} & \lightblue{\textbf{37.2309}} & 32.0777 & 31.2692 & 33.1866 & 35.4194 & \cellcolor{mygray}\blue{\textbf{37.0989}} \\
PET $\to$ CT & 0.9161 & 0.9148 & 0.9215 & 0.9070 & 0.9121 & \cellcolor{mygray}\blue{\textbf{0.9282}} & 28.1046 & 29.3181 & \lightblue{\textbf{29.6694}} & 26.8885 & 27.2708 & \cellcolor{mygray}\blue{\textbf{30.1548}} \\
CT $\to$ PET & \blue{\textbf{0.9884}} & 0.9824 & 0.9851 & 0.9834 & 0.9842 & \cellcolor{mygray}\lightblue{~\textbf{0.9883}~} & \lightblue{\textbf{39.8490}} & 39.0795 & 39.1718 & 39.4528 & 39.4348 & \cellcolor{mygray}\blue{\textbf{41.5840}} \\ \hline
Avg. & \lightblue{\textbf{0.9672}} & 0.9571 & 0.9610 & 0.9568 & 0.9646 & \cellcolor{mygray}\blue{\textbf{0.9728}} & 37.5164 & 35.6744 & 36.6677 & 36.2777 & \lightblue{\textbf{37.3862}} & \cellcolor{mygray}\blue{\textbf{39.5963}} \\ \hline
\end{tabular}%
}
\captionof{table}{\textbf{Ablation study - Modality transfer results of PET and CT} on AutoPET-II: Ablation results of models trained under different combinations of constraints. 
The best and second results are highlighted in \blue{\textbf{blue}} and \lightblue{\textbf{cyan}}, respectively. 
}
\label{tab:gen_AutoPET-II_main}
\end{minipage}%
\hfill 
\begin{minipage}{0.18\linewidth}
    \centering
    \vspace{-0.7cm} \includegraphics[width=0.8\linewidth]{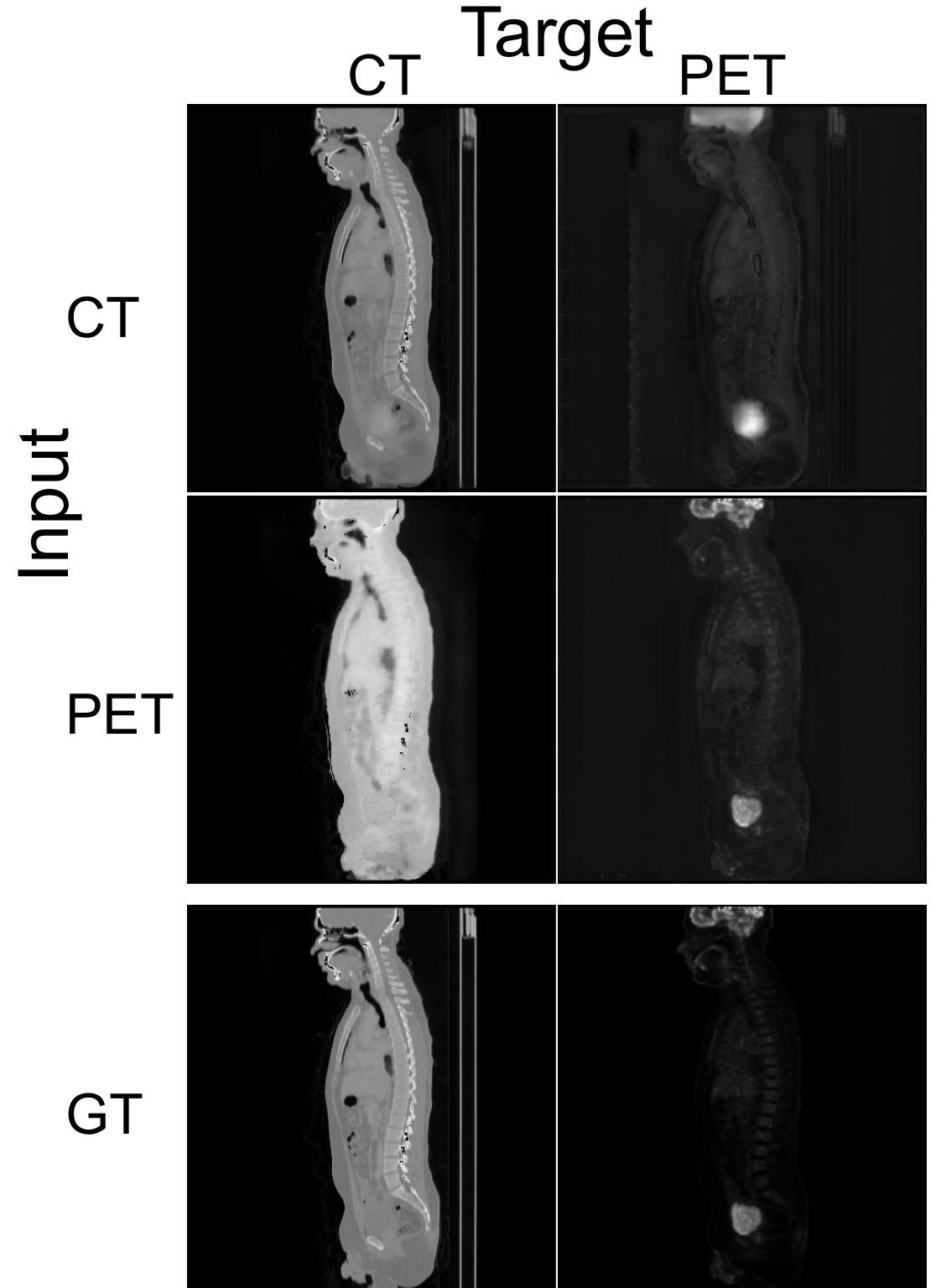}
\caption{Generated samples on AUTOPET-II.}
\label{fig:autpet}
\vspace{-0.3cm}
\end{minipage}
\vspace{-0.4cm}
\end{figure*}

\textbf{Results.}
\cref{tab:gen_BRATS23_highlight} exhibits the transfer results in comparison between our and previous one-to-one modality transfer methods.
Following previous 
of T1 $\to$ T2 and T2 $\to$ Flair. Our approach significantly surpasses previous 2D and 3D generation methods, including single- and multi-modality translation methods.  
Specifically, our approach exceeds current SOTA diffusion-based methods, such as 2D-based MD-Diff and 3D-based MS-SPADE.
In terms of one-to-all multi-modality translation, 
as \cref{tab:gen_BRATS23_main} and additional results in the Appendix show, our approach with various volume sizes ($96$ and $192$) performs better than MS-SPADE and SwinUNETR across all metrics under all settings.
Moreover, our method significantly improves the SSIM, indicating a better anatomy structure obtained by our approach. 
These results indicate that the $\mathbb{X}_h$ Hypothesis is plausible for homogeneous generalization, and our personalized approach is able to obtain its approximation. 
Please refer to more analysis on $\mathbb{O}$ in \cref{app:prior}.


\begin{figure}[t]
         \centering
        \vspace{-0.1cm}
        \includegraphics[width=0.95\linewidth]{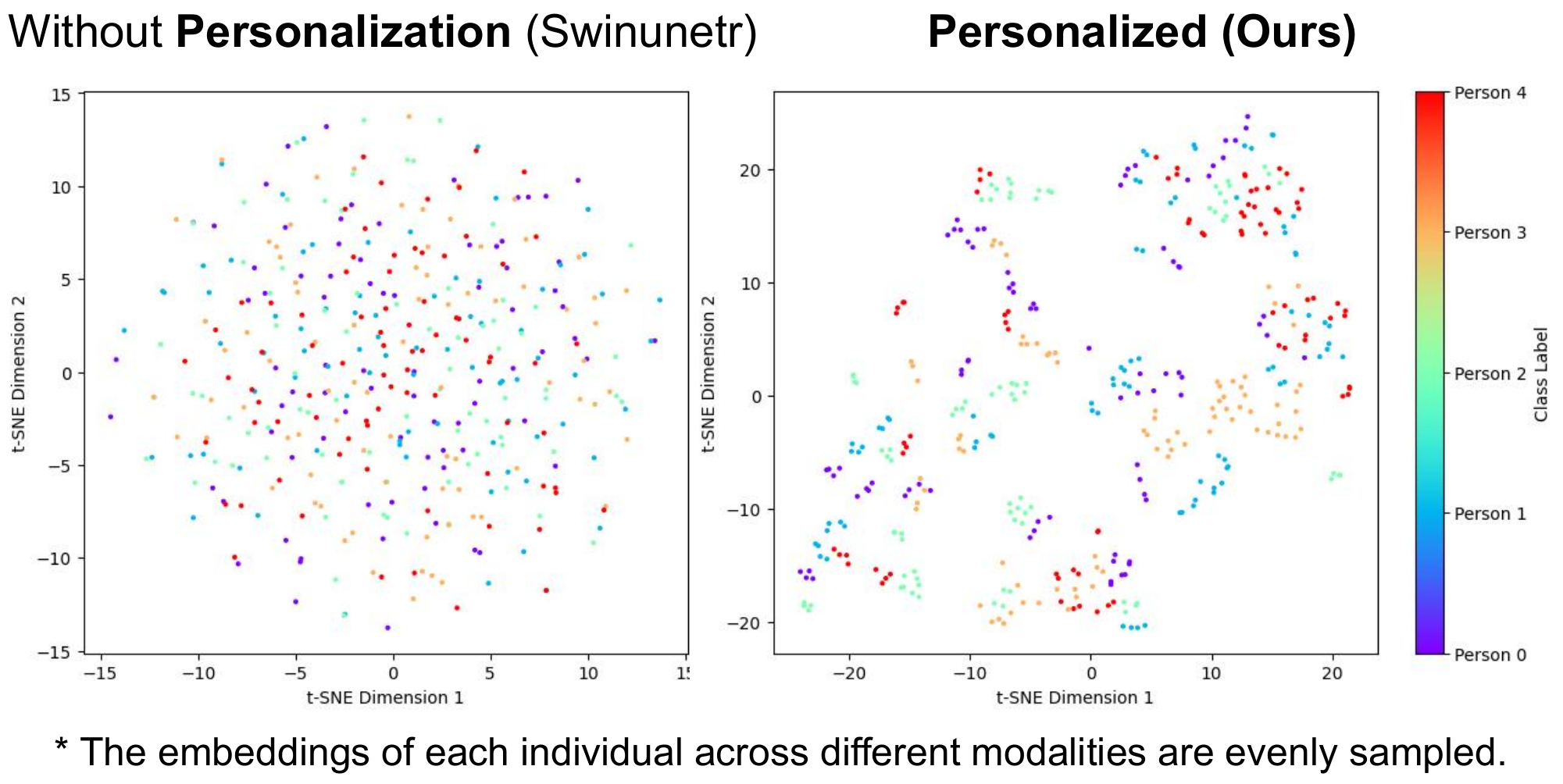} 
        \vspace{-0.2cm}
        \caption{T-SNE visualizations on embeddings of SwinUnetr and ours trained on BRATS23 for one-to-all modalities generation.}
\label{fig:tsne_of_personalization}
\vspace{-0.3cm}
\end{figure}

\textbf{More analysis on personalization.}
\cref{fig:tsne_of_personalization} also shows that combining the prior and constraints results in more scattered embeddings at the individual level compared to the method without them. \cref{tab:compare SSL}, \cref{tab:missing_seg_brats18_main}, and the quantitative results in the paper further confirm the advantages of personalization.

\subsubsection{Fine-tuning stage: Missing modality segmentation}

\textbf{Experimental settings.} 
To validate the generalization ability of the pre-trained model, we fine-tune the model obtained from \cref{sec:seq_mod_trans} on the BRATS18~\citep{menze2014multimodal} from the Multimodal Brain Tumor Segmentation Challenge.
Similar to BRATS23, BRATS18 also consists of the same four structural modalities. We employ the Dice similarity
coefficient (DICE) as the metric for evaluation. 
For a fair comparison, we follow data splits of \cite{shi2023m} and reproduce the results of {previous methods~\citep{ding2021rfnet,zhang2022mmformer,wang2023multi,liu2023m3ae,shi2023m} }on these splits by using their released code and following their original settings~\footnote{Though we tried our best, it can be noticed some reproduced results are lowered than their reported results in their original paper. It should be clarified that \textbf{our results also exceed those reported results}.
However, for a comprehensive study, we mainly report our reproduced results. 
}. 
{We also reproduce the results by using the SwinUNETR pre-training method~\citep{tang2022self}, employing their official code to train the model on the same datasets with identical training epochs, learning rates, and other hyperparameters as ours.}
See additional experimental details in \cref{app:exp}.




\textbf{Results.}
\cref{tab:missing_seg_brats18_main} presents the segmentation results of our approach compared to previous methods. We also compute the standard deviation of DICE scores under various missing modality settings, which highlights the robustness of models. Notably, our approach outperforms previous methods in most missing modality scenarios, particularly when the number of missing modalities is large (e.g., NM=3).
Meanwhile, our approach also obtain a relatively low standard deviation for most settings, showing its robustness to various modalities missing scenario.
{Compared with SwinUNETR pre-training, our methods yield consistently better DICE with comparable standard deviation.} 
This performance improvement stems from the enhanced generalization of our model, which is rooted in the learned ${\mathbb{X}_h}'$.

\subsection{Heterogeneous generalization: PET and CT}
\label{sec:heterogeneous_modalities}

\subsubsection{Pre-training stage: Modality transfer}
\label{sec:pretrain_hetero_transfer}

\textbf{Experimental settings.}
We utilize the AutoPET-II dataset from the Automated Lesion Segmentation in PET/CT challenge~\cite{gatidis2022fdg} for pre-training. The AutoPET-II dataset includes AC-PET and CT pairs, where the PET scans adopt FDG tracers, and their attenuation is corrected using the corresponding CT scans.
Specifically, we divide the AutoPET-II dataset into training and testing sets. Similar to our approach for heterogeneous generalization, we adopt the Peak Signal-to-Noise Ratio (PSNR) and Structural Similarity Index (SSIM) as evaluation metrics.
In this section, we present the results of models that employ different combinations of the constraints and the set $\mathbb{O}$. We use both contrastive loss and the decomposition constraint as our baseline. {Please refer to training details in \cref{app:exp}.}

\begin{table*}[t!]
\begin{minipage}{0.59\linewidth}
\centering
\resizebox{\textwidth}{!}{%
\begin{tabular}{l|ccc|ccc}
\hline
\multicolumn{1}{c}{} & \multicolumn{3}{c|}{SSIM↑} & \multicolumn{3}{c}{PSNR↑} \\ \hline
HNSCC validation &  NAC→CT & NAC →AC & Avg. & NAC→CT & NAC→AC  & Avg.\\ \hline
UNETR & 0.4899& 0.8998& 0.6949	& 21.7330 & 42.8557 &  32.2944\\
SwinUNETR &0.5853 &0.9265 & 0.7559& 23.5628& 42.5495&33.0561 \\
\rowcolor{mygray}\textbf{Ours (PUIR)} & \blue{\textbf{0.6939}} & \blue{\textbf{0.9516}} & \blue{\textbf{0.8227}} & \blue{\textbf{25.8498}} & \blue{\textbf{46.4658}} & \blue{\textbf{36.1578}}						

\\ 
\hline
\multicolumn{1}{c}{} & \multicolumn{3}{c|}{SSIM↑} & \multicolumn{3}{c}{PSNR↑} \\ \hline
NSCLC &  NAC→CT & NAC →AC & Avg. & NAC→CT & NAC→AC  & Avg.\\ \hline
UNETR & 0.4476 & 0.8703 & 0.6590 & 20.6182 & 40.8570 & 30.7376 \\
SwinUNETR & 0.4476 & 0.8705 & 0.6591 & 22.5086 & {{41.3272}}  & 31.9179\\
\rowcolor{mygray}\textbf{Ours (PUIR)} & \blue{\textbf{0.4744}} & \blue{\textbf{0.8853}} & \blue{\textbf{0.6798}} & \blue{\textbf{22.7791}} & \blue{\textbf{42.7687}} & \blue{\textbf{32.7739}}				
\\ 
\hline
\end{tabular}%
}
\captionof{table}{\textbf{Modality transfer results of NAC-PET to AC-PET and CT} that tuned on HNSCC and evaluated on HNSCC validation set and NSCLC: 
Comparison between the previous method and ours for transfer between different modalities.
The best results are highlighted in \blue{blue}. 
}
\label{tab:gen_nsclc_main}
\end{minipage}
\hfill
\begin{minipage}{0.39\linewidth}
\centering
\includegraphics[width=\linewidth]{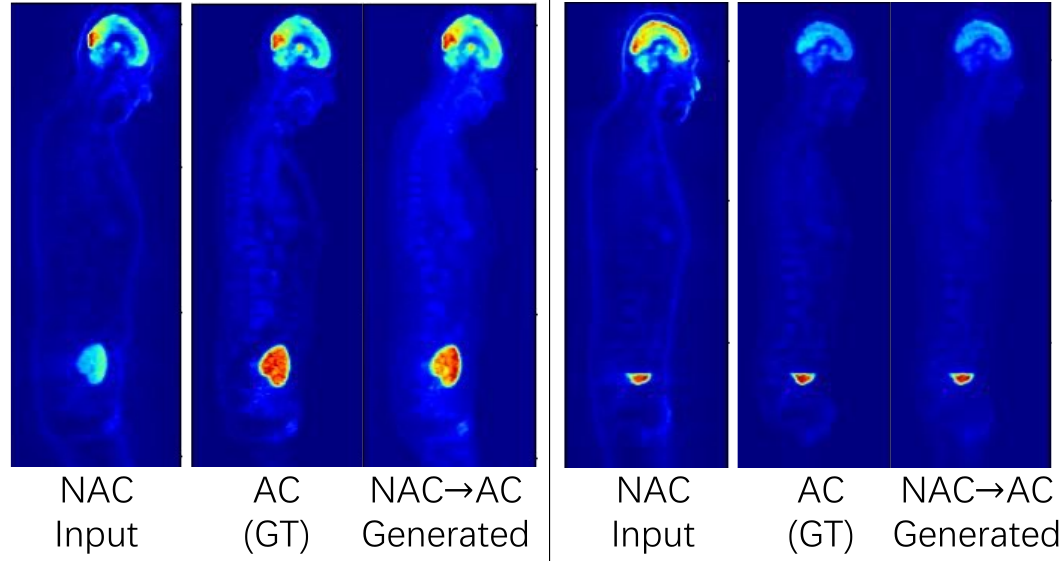}
\vspace{-0.5cm}
\captionof{figure}{\textbf{Modality transfer results of NAC-PET to AC-PET:} Generated examples on the NSCLC dataset for NAC $\to$ AC across individuals.}
\label{fig:gen_nsclc_main}
\end{minipage}
\vspace{-0.4cm}
\end{table*}


\textbf{Results.} 
As shown in \cref{tab:gen_AutoPET-II_main} and generated examples in \cref{fig:autpet}, incorporating $\mathbb{O}$ with different combinations of constraints improves generation quality across most metrics. Specifically, using the constraints without $\mathbb{O}$ does not guarantee improvements, as discussed in \cref{sec:contraints}.
Ultimately, employing all constraints along with $\mathbb{O}$ yields the best average results across all translations, validating that our approach performs well in heterogeneous generalization settings. 
These results indicate that our method under the scope of personalization bridges the gap between structural and functional modalities.  
{We validate the transferability of all these pre-trained models in 
\cref{app:seg_ablation},  where additional analyses are provided.}




\subsubsection{Fine-tuning stage: Segmentation}

\textbf{Experimental settings.} 
We utilize the AutoPET-II~\cite{gatidis2022fdg} dataset for segmentation, evaluating performance using the DICE metric. It is important to note that we employ the same training and testing splits as in \cref{sec:pretrain_hetero_transfer} to avoid data leakage. Specifically, we adhere to the settings from the official challenge; DICE is calculated in the standard manner but is set to zero for false negatives and true negatives. Additionally, we introduce DICE- to include the mean across all samples, along with true positive rate (TPR), true negative rate (TNR), false negative rate (FNR), and false positive rate (FPR) for the missing modality segmentation evaluation. Our method is compared against nnUNET~\citep{isensee2021nnu}, UNETR~\citep{hatamizadeh2022unetr}, and SwinUNETR~\citep{hatamizadeh2022swin}, which are trained directly on the dataset without pre-training. Notably, we also compare our approach with SwinUNETR using its pre-training strategy~\citep{tang2022self}.
{Please refer to training details in \cref{app:exp}.}

\begin{wraptable}{r}
{0.42\linewidth} 
    \centering
    \vspace{-0.2cm}
    \resizebox{\linewidth}{!}{%
        \begin{tabular}{lc}
        \hline
        From scratch & Dice \\ \hline
        nnUnet~\citep{isensee2021nnu} & 33.1 \\
        SwinUNETR~\citep{hatamizadeh2022swin} & 43.5 \\ \hline
        \multicolumn{2}{l}{With 3D medical image SSL} \\ \hline
        SwinUNETR~\citep{tang2022self}  & 44.1 \\
        SwinMM~\cite{wang2023swinmm} & 44.2 \\
        PCRL v2~\cite{zhou2023pcrlv2} & 41.9\\
        VoCo~\cite{wu2024voco} & 46.3 \\
        \rowcolor{mygray}\textbf{Ours (PUIR)} & \blue{\textbf{48.2}} \\ \hline
        \end{tabular}%
        }
        \vspace{-0.3cm}
        \captionof{table}{Comparison between different SOTA medical 3D SSL methods on AUTOPET segmentation.}
        \label{tab:compare SSL}
        \vspace{-0.3cm}
\end{wraptable}
\textbf{Results.}
{The segmentation results are presented in \cref{tab:compare SSL}, with additional details provided in Appendix~\cref{tab:seg_AutoPET-II}. For a fair comparison, all other SSL methods were reproduced using the same pre-training and fine-tuning datasets as those used in our approach.  Our approach significantly improves the DICE results in comparison to other SSL methods since we specifically address the cross-modality generalization. 
The performance gains on heterogeneous modalities also enhance the significance of our $\mathbb{X}_h$ hypothesis for heterogeneous generalization.}


\subsection{Fine-tuning special case: A complex scenario}
\label{sec:special_case}

We introduce a more complex scenario in which the pre-trained model for heterogeneous generalization settings is tuned downstream to span both heterogeneous and homogeneous generalization. 

\textbf{Experimental settings.} 
The pre-train model we adopted is from \cref{sec:heterogeneous_modalities} that trained on AC-PET and CT. 
Specifically, we tune the model by using the Head and Neck Squamous Cell Carcinoma (HNCSS) dataset~\citep{grossberg2020md} as the training set and the Non-Small Cell Lung Cancer (NSCLC) dataset as the testing set. Both datasets are sourced from The Cancer Imaging Archive (TCIA)~\citep{clark2013cancer}, and they contain paired non-attenuation-corrected PET (NAC-PET), attenuation-corrected PET (AC-PET), and CT scans.
The model is pre-trained for heterogeneous generalization between AC-PET and CT. It is tuned for both homogeneous generalization between AC-PET and NAC-PET and heterogeneous generalization between NAC-PET and CT. 
Similar to the previous translation experiments, we use SSIM and PSNR as evaluation metrics. Performance in this scenario further validates the model’s generalization capabilities.
Note here the training and testing data in the downstream task come from different domains.
{See training details in \cref{app:exp}.}






\textbf{Results.}
\cref{tab:gen_nsclc_main} presents the results on the HNSCC dataset, while \cref{fig:gen_nsclc_main} displays generated sample images for homogeneous generalization. 
Our approach achieves superior results across both heterogeneous and homogeneous generalizations.
For heterogeneous generalization, our method consistently improves SSIM for NAC-PET to CT, indicating that the learned ${\mathbb{X}_h}'$ successfully captures and emphasizes anatomical structures in the generated images, as indicated by improved SSIM. 
Moreover, though the model is pre-trained between AC-PET and CT, the improvements are also consistent for NAC-PET and AC-PET.
These findings confirm that our personalized approach is effective for a complex real-world scenario, demonstrating the transferability and generalizability of the pre-trained model to downstream tasks under various scenarios.

\section{Conclusion}
This paper proposes a universal approach to address multi-modality generalization by approximating a personalized invariant representation, $\mathbb{X}_h$, through constraints of invariance, equivariance, and decomposition, guided by a learnable biological prior. We demonstrate that learning $\mathbb{X}_h$ is both feasible and highly beneficial for enhancing generalization in medical tasks. 
Our method also correlates to personalized medicine, a transformative framework for $21^{st}$ century healthcare, tailoring medical treatments to each patient's unique characteristics~\citep{whitcomb2012personalized,katsanis2008case,chan2011personalized}.
By discussing limitations, future directions, and social impact in \cref{app:limitations,app:social impact}, our study may point to a promising path for achieving medical generalization through personalization in complex multi-modality medical analysis.

\section*{Acknowledgments}
This work was supported by the National Natural Science Foundation of China (Grant No. 92370119, No. 62376113, No. 62206225, No. 62436009, No. 62276258, No. 82394432, and No.92249302), the Shanghai Municipal Science and Technology Major Project (Grant No. 2023SHZDZX02), and XJTLU Funding REF-22-01-002. 
The computations in this research were performed
using the CFFF platform of Fudan University.


{
    \small
    \bibliographystyle{ieeenat_fullname}
    \bibliography{main}
}


\clearpage
\setcounter{page}{1}
\maketitlesupplementary
\appendix
\section{Brief descriptions of different modalities}
\label{app:modalities}

\textbf{Magnetic Resonance Imaging (MRI) scans}~\citep{zhao2022modality}: use strong magnetic fields and radiofrequency currents yielding distinct sequences.
Typically, MRI has different modalities, include T1, T2, T1ce and Flair. 

\textbf{Computed Tomography (CT) scans}~\citep{ozbey2023unsupervised,zhan2024medm2g} employ X-rays to measure its attenuation. 

\textbf{Positron Emission Tomography (PET) scans} are expensive functional imaging scans that employ radiotracers emitting gamma rays to visualize and measure metabolic processes. Thus, PET scans have a large percentage of background areas.

\section{Related work}
\label{app:Related work}


\textbf{Medical generalization tasks.} Most current work focuses on homogeneous generalization, introducing tasks such as modality transfer and missing modality segmentation. The most commonly employed structural modalities — Flair, T1, T2, and T1ce of MRI — are used for brain tumor segmentation~\citep{zhao2022modality}, or between MRI and CT~\citep{zhan2024medm2g} for modality transfer.  \cite{pan2023revealing} propose an approach for heterogeneous generalization in terms of modality transfer, but only tailored for transferring PET to CT.

\textbf{Self-supervised medical pre-train models for medical generalization.}
Our approach aims to learn the $\mathbb{X}_h$ through pre-training. We list related medical pre-training work 
\cite{tang2022self,wu2024voco,chen2020mocov2,jiang2023anatomical} here. A notable work among them is \cite{jiang2023anatomical}, which extracts class-specific anatomical invariance. However, they only focus on a single modality. Such single-modality approaches may not be able to construct $\mathbb{X}_h$ for improving the generalization across modalities.

\textbf{Alignment in multi-domain generalization.}
The issue of cross-modality generalization is similar to the problem of multi-domain generalization, which aims to extract domain invariant representations~\citep{ganin2016domain,li2018deep,li2018domain,hu2020domain,tan2024rethinking}. Most of these approaches focus on learning invariance across different domains, which may not fit the scope of personalization. 


\textbf{Generalization for medical translation.}
Typical modality transfer approaches are based on GAN models~\citep{isola2017image,zhu2017unpaired,fu2019geometry,park2020contrastive,kong2021breaking}. In contrast to these GAN-based approaches, some work adopts transformer models~\citep{liu2023one,shi2023m}, while others, such as \cite{dhariwal2021diffusion,ozbey2023unsupervised,kim2024adaptive,xing2024cross}, explore diffusion-based approaches.
The methods such as MedM2G~\citep{zhan2024medm2g} further incorporate textual information for modality transfer. 
Additionally, UNET-like architectures, which can also be applied to these tasks, are highlighted in~\citep{hatamizadeh2022unetr,hatamizadeh2022swin}.
Most current modality transfer research focuses on improving synthesis quality. Our approach, however, demonstrates that full-modality transfer, when accompanied by specific constraints, not only enhances generation but also improves downstream generalization.

\textbf{Generalization for medical segmentation.}
There are three main types of approaches to missing modality segmentation.
Knowledge distillation-based approaches transfer knowledge from models with complete modality information (teachers) to models with missing modality information (students)~\citep{chen2021learning,wang2023prototype}. 
\citep{ding2021rfnet,zhang2022mmformer} recover missing information by leveraging the multimodal latent feature space. 
Domain adaptation-based methods aim to reduce the gap between models with complete and incomplete modalities by aligning their domains~\citep{wang2021acn}.
One prominent shared latent space method, MmFormer~\citep{zhang2022mmformer}, exploits intra- and inter-modality dependencies for feature fusion, which is closely related to our work. Our work reveals that our pre-train model with basic segmentation tuning exceeds these approaches.

\section{Limitations, challenges, and future work}
\label{app:limitations}
To enhance the validation of our approach, we adhere to commonly used settings during the tuning stage. Exploring alternative strategies, such as knowledge distillation, could further improve downstream performance.
Our approach requires datasets where all modalities are instance-level matched, which can be a stringent condition and may be unattainable for certain modalities. Future research should explore methods to achieve personalized invariance without relying on instance-level matched datasets. 
Additionally, we advocate for the availability of more open-source multi-modal medical datasets, particularly for functional modalities, as these are not widely accessible to researchers.

\section{Social impact}
\label{app:social impact}
This work presents an approach to tackle multi-modality generalization through personalization. We hope our work can encourage the community to work towards practical, personalized medical models with border generalization ability.

\section{Downstream segmentation ablation study}
\label{app:seg_ablation}

\begin{table*}[ht]
\caption{\textbf{Ablation study - Segmentation results of using different pre-train models} on AutoPET-II: Comparison between the pre-train models with different settings and ours. 
The best results are highlighted in \blue{blue} and \lightblue{cyan}. 
} 
\label{tab:ablation_res}
\resizebox{\textwidth}{!}{%
\begin{tabular}{cl|c|ccccc}
\hline
  ID & Pretrian & \textbf{DICE}$\uparrow$ & DICE-$\uparrow$ & TPR$\uparrow$ & TNR$\uparrow$ & FNR$\downarrow$ & FPR$\downarrow$ \\ \hline
1 & + Contrastive + Decomposition + Equivariance & 40.85 & 55.79 & 81.72  & 69.09 & 18.28  & 30.91 \\
2 & + Contrastive + Decomposition + Invariance   &  44.34 & 48.63 & 77.42 & 91.82 & 22.58 & \blue{\textbf{8.18}} \\
3 & + Contrastive + Decomposition + Equivariance + Invariance & 42.42 & 60.67 & \blue{\textbf{89.25}} & 63.64 & \blue{\textbf{10.75}}  & 36.36 \\ \hline
4 & + Contrastive  + Decomposition + Equivariance + $\mathbb{O}$ &  46.31 & 55.77 & 83.87 & 82.73 & 16.13  & \lightblue{17.27} \\
5 & + Contrastive  + Decomposition + Invariance + $\mathbb{O}$ & 44.42  & 57.80 & \lightblue{88.17} & 74.55 & 11.83 &  25.45\\
6 & \cellcolor{mygray}\begin{tabular}[c]{@{}l@{}} \textbf{+ Contrastive + Decomposition + Equivariance + Invariance + $\mathbb{O}$}\end{tabular} &
\cellcolor{mygray}\blue{\textbf{48.20}} & \cellcolor{mygray}\blue{\textbf{61.16}} & \cellcolor{mygray}{\lightblue{\textbf{88.17}}} & \cellcolor{mygray}\blue{\textbf{77.27}} & \cellcolor{mygray}\lightblue{\textbf{11.83}} & \cellcolor{mygray}{{22.72}}
\\ \hline
\end{tabular}%
}
\end{table*}

The effectiveness of our proposed components is demonstrated alongside an exploration of the methodology employed to develop an individual-invariant representation. 
Experimental results for downstream segmentation tasks and visualizations of the pre-trained models are presented in \cref{tab:ablation_res}. 
All experiments are conducted under consistent settings to ensure a fair comparison.


\textbf{Using all constraints together with $\mathbb{O}$ yields the best results.} 
Consistent with \cref{sec:contraints}, the results indicate that using different constraints alone may not guarantee improvements; however, incorporating all constraints along with $\mathbb{O}$ results in the best outcomes. This validates the plausibility of the $\mathbb{X}_h$ Hypothesis and demonstrates that achieving a good approximation of it significantly enhances generalization.

\textbf{Using prior $\mathbb{O}$ with decomposition constraint improves the model performance for different settings.} 
Despite different settings, additionally using $\mathbb{O}$ with decomposition improves the downstream model performance.
Combined with the improvements from modality transfer results in \cref{tab:gen_AutoPET-II_main}, it suggests that $\mathbb{O}$ helps in better obtaining anatomical structure.

\textbf{The invariance and equivariance constraints can not be applied to the same feature}. It needs to be highlighted that invariance and equivariance constraints cannot be applied to the same features as they conflict with each other. As shown in task 3, without $\mathbb{O}$, invariance and equivariance constraints are applied to the latent feature simultaneously, leading to a significant performance drop. In comparison, applying equivariance constraint before using $\mathbb{O}$ and applying the invariance constraint after using $\mathbb{O}$ yields the best results. This is because the geometrical transformation contained in $z^i_h$ needs to be accomplished by fetching other possible geometrical transformation information from $\mathbb{O}$ and then fusing it to be invariant.









\section{Experimental details}
\label{app:exp}

The model and data loaders are built by using MONAI~\url{https://docs.monai.io/en/stable/index.html}. 
Please refer to all the details of the implementation in the code. We present some key implementations below.

\subsection{Overall training procedure}
A pseudo-code is provided for our approach. The loss calculation for 
\textbf{Pre-training} procedure is simplified as Algorithm~\ref{alg:pre-training} and \textbf{Downstream tuning} as Algorithm~\ref{alg:fine-tuning}. 
It is notable that the empirical procedure is flexible as long as the $\mathbb{O}$ is properly used to construct $\mathbb{X}_h'$ and those constraints are applied to $\mathbb{X}_h'$. 

\RestyleAlgo{ruled}
\SetKwComment{Comment}{/* }{ */}
\begin{algorithm}[hbt!]
\caption{Calculate losses during one step for pre-training}
\label{alg:pre-training}
\KwData{$X\in \mathcal{X}, epoch$}
Initialize learnable $\mathbb{O}$ $\mathcal{E}(\cdot), \mathcal{D}(\cdot)$\;
\While{$i \neq epoch$}{
  $X_h' \gets None$\;
  \For{$h \in \mathcal{H}$}{
  \For{$i \in \mathcal{M}$}{
  $\mathcal{L}_{pre} \gets 0$\;
  {$X_h^i \sim X$, $\phi^i \sim \Phi$\;
  ${X_h^i}^+, {X_h^i}^- = Augment(\phi^i(X_h^i))$}\;
  $({z_h^i}, {x_h^i}), ({z_h^i}^-,{x_h^i}^-), ({z_h^i}^+,{x_h^i}^+) \gets \mathcal{E}(\phi^i(X_h^i)), \mathcal{E}({X_h^i}^-), \mathcal{E}({X_h^i}^+)$\;
  \blue{Calculate $\mathcal{L}_{contr}({z_h^i}, {z_h^i}^+, {z_h^i}^-) $, $\mathcal{L}_{pre} += \mathcal{L}_{contr}$}\;
  $\mathcal{F}(z_h^i) \to {\phi^i}'$\;
  \blue{Calculate $\mathcal{L}_{equ}({\phi^i}', {\phi^i}) $, $\mathcal{L}_{pre} += \mathcal{L}_{equ}$}\;
   ${z^i_h}' := Attn(query:z^i_h, key:\mathbb{O}, value:\mathbb{O})$\;
  ${X_h^i}' := Conv(z^i_h{}',z^i_h)$ \;
  \eIf{$X_h'$ is not None \Comment*[r]{For saving memory}} 
    {
       \blue{Calculate $\mathcal{L}_{inv}(X_h^i{}', X_h')$,  $\mathcal{L}_{pre} += \mathcal{L}_{inv}$}\;
        $X_h' := (X_h+X_h^i{}')/2$\;
        
    }{
        $X_h':= X_h^i{}'$ \;
    }
    $X_h^i{}' := \mathcal{D}( X_h^i{}', x_h^i)$\;
    \blue{Calculate $\mathcal{L}_{decom}(\phi^i{}^{-1}(X_h^i{}'), X_h)$,  $\mathcal{L}_{pre} += \mathcal{L}_{decom}$}\; 
}
}
}
\end{algorithm}

\begin{algorithm}[hbt!]
\caption{Calculate losses during one step for fine-tuning}
\label{alg:fine-tuning}
\KwData{$(X, Y )\in (\mathcal{X}, \mathcal{Y}), epoch$}
Load pre-trained $\mathbb{O}$ $\mathcal{E}(\cdot), \mathcal{D}(\cdot)$\;
\While{$i \neq epoch$}{
  $X_h' \gets None$\;
  \For{$h \in \mathcal{H}$}{
  \For{$i \in \mathcal{M}$}{
  $\mathcal{L}_{down} \gets 0$\;
  $(X_h^i, Y_h)\sim X, Y$ \;
    $({z_h^i}, {x_h^i}) \gets \mathcal{E}({X_h^i})$\;
   ${z^i_h}' := Attn(query:z^i_h, key:\mathbb{O}, value:\mathbb{O})$\;
  ${X_h^i}' := Conv(z^i_h{}',z^i_h)$ \;
  \eIf{$X_h'$ is not None \Comment*[r]{For saving memory}}{
        \blue{Calculate $\mathcal{L}_{inv}(X_h^i{}', X_h')$,  $\mathcal{L}_{down} += \mathcal{L}_{inv}$}\;
        $X_h' := (X_h+X_h^i{}')/2$\;
    }{
        $X_h':= X_h^i{}'$ \;
    }
    $Y_h' := \mathcal{D}( X_h^i{}', x_h^i)$\;
    \blue{Calculate $\mathcal{L}_{ori}(Y_h', Y_h)$, $\mathcal{L}_{down} += \mathcal{L}_{ori}$}\;
}
}
}
\end{algorithm}


\subsection{Homogeneous generalization: structural modalities in MRI}

\subsubsection{Pre-training and Modality transfer.} 
\textbf{Experimental settings.} 
Four A100 GPUs are employed for training. 
The learning rate we used for the modality transfer is set to $0.0002$, and the training epoch is set to $1000$. 
Both the number of input and out channels is set as $4$.  

\textbf{Training details.} 
For the model, both the input and output channels are set to $4$, corresponding to the four MRI modalities. All modalities are loaded and cropped to a size of $96\times96\times96$ simultaneously. Following~\citep{kim2024adaptive}, we also normalize each MRI modality to have zero mean and unit variance. During training, the background is excluded for modal generation. 
A single modality is repeated four times to create four channels during training to obtain ${\mathbb{X}^i_h}'$.
The training loss follows the $\mathcal{L}_{pre}$, whose calculation details during the training phase can be seen in Algorithm~\ref{alg:pre-training}.

\subsubsection{Missing modality segmentation.} 
Four A100 GPUs are employed for tuning. 
The learning rate we used for the modality transfer is set to $0.0002$, and the training epoch is set to $1000$. 
Both the number of input and out channels is set as $4$.  

\textbf{Training details.} 
Following~\cite{shi2023m}, we also normalize each MRI modality to zero mean and unit variance. 
For the fine-tuning, we 
employ Dice loss, the weighted
cross-entropy loss that is adopted by \cite{shi2023m}, and the additional $\mathcal{L}_{inv}$.

\subsection{Heterogeneous generalization: PET and CT modalities}

\subsubsection{Modality transfer}
All models are trained using A100 GPUs.
\textbf{Training details.} All models are trained under the same situations, using the same data pre-processing transforms.

\subsubsection{Downstream segmentation}
\textbf{Training details.} 
All training and fine-tuning experiments use the same losses, while the approaches with our pre-train additionally use $\mathcal{L}_{inv}$ for downstream fine-tuning.
Moreover, we also compare the original architecture of SwinUNETR using our pre-trained weights with fully using our architecture and our weights for fine-tuning.  


\subsection{Fine-tuning special case: Tuning from heterogeneous to homogeneous generalization with domain gap}
\textbf{Training details.} 
For the fine-tuning stage, we use the decoder architecture of SwinUNETR, which is randomly initialized. The training procedure is similar to the above modality transfer experiments, with the primary difference being that the input and output channels are set to two. Additionally, we reproduced the results of UNETR and SwinUNETR for comparison, ensuring that the same loss functions were applied across models.

\section{More results and analysis}
\label{app:more_res}



\subsection{More analysis on learnable biological prior}
\label{app:prior}
\begin{figure*}[t]
\begin{center}
\includegraphics[width=\linewidth]{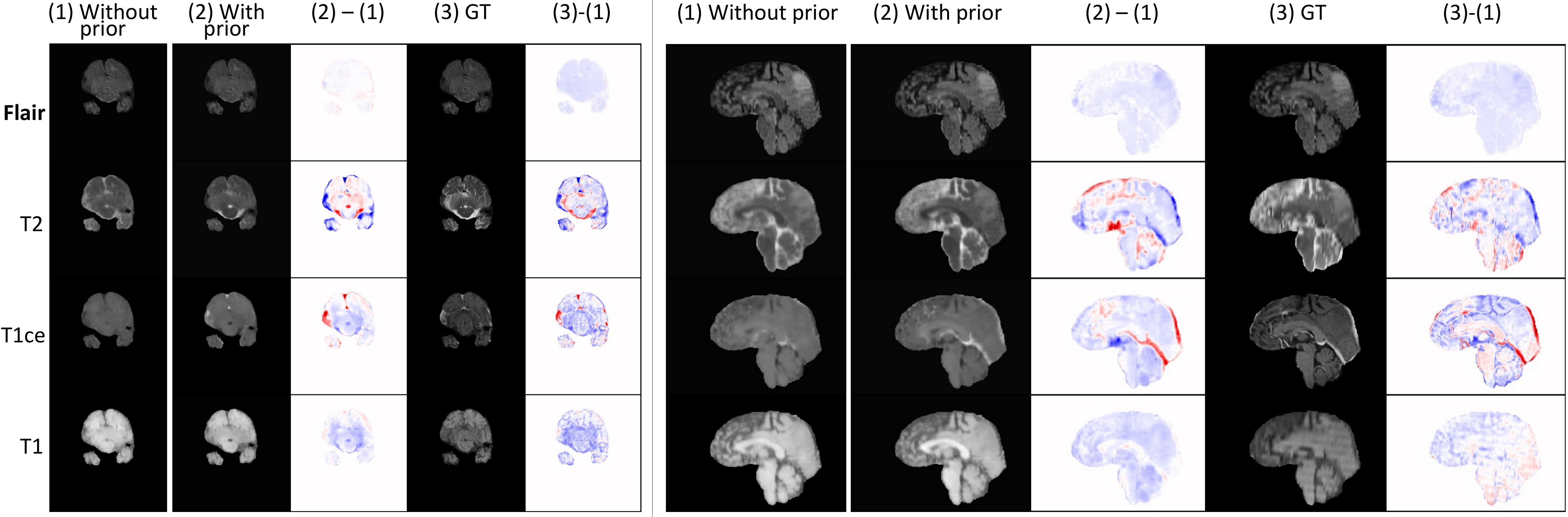}
\end{center}
\vspace{-0.5cm}
\caption{Visualization of the efficacy of  prior $\mathbb{O}$. 
Displayed are the generated modalities on the input Flair modality of a testing sample on the BTATS21 dataset. Columns show: the generated images of the model (1) without prior $\mathbb{O}$ and (2) with prior $\mathbb{O}$ are aligned with (3) the GT images. 
Typically, the differences between without and with prior   $\mathbb{O}$ (the (2)-(1) column) are visualized to compare with the differences between without $\mathbb{O}$ and GT (the (3)-(1) column). \red{Red} and \blue{blue} refer to the positive (accomplishment) and negative (refinement) values of the differences, respectively.
}
\label{fig:Prior}
\vspace{-0.3cm}
\end{figure*}

\textbf{Analysis of $\mathbb{O}$.}
We show that using $\mathbb{O}$ for $\mathbb{X}_h$ mainly accomplishes the personalized knowledge of each sequence from MRI modalities. 
Those modalities are mainly focused on the physical anatomy. 
For the Flair modality in MRI, which mainly highlights the lesion but suppresses structures like bones, \cref{fig:Prior} shows that without $\mathbb{O}$, the main difference between the generated images and ground truth (GT) images is the personalized structure. 
Prior $\mathbb{O}$ for $\mathbb{X}_h$ accomplishes and refines the personal level anatomical information, mitigating the gap between them with the GT, so it can be better transferred to other structural focusing modalities.


\begin{figure*}
    \includegraphics[width=\linewidth]{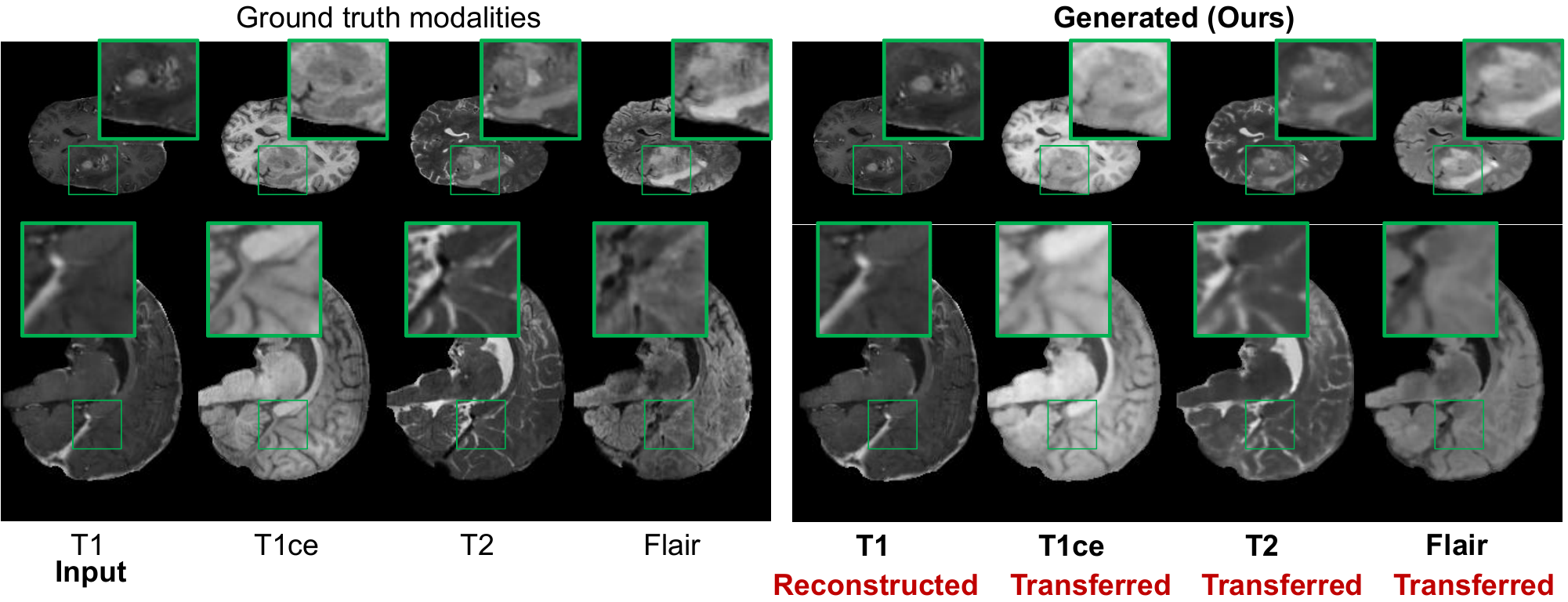} 
        \vspace{-0.2cm}
        \caption{
        Visualizations of generated modalities with T1 as input of our method, which allows the capturing of subtle structures.
        }
        \label{fig:vis_zoomed_in}
\end{figure*}

\subsection{Segmentation results on  AutoPET-II.}
Detailed metrics results of  AutoPET-II are presented in \cref{tab:seg_AutoPET-II}.
The results indicate that with proper model architecture, such as SwinUNETR, using both two modalities usually outperforms solely using PET. 
It can be observed that models using our pre-train improve the results across all metrics.
Typically, SwinUNETR using our pre-train significantly exceeds it without our pre-trained model, indicating the personalized invariant learned by our pre-train generalizes to the downstream well and can boost the downstream tasks.
Moreover, using our proposed components with the pre-train leads to the best DICE and DICE-. This validates that using the prior further emphasizes the personalized invariant, which yields the most segmentation improvements. 

\begin{table*}[!t]
\centering
\begin{tabular}{l|c|ccccc}
\hline
 Method  & \textbf{Dice↑} & Dice-↑ & TPR↑ & TNR↑ & FNR↓ & FPR↓ \\ 
 \hline
\multicolumn{1}{c}{} & \multicolumn{6}{c}{From scratch}
\\ \hline
nnUnet~\citep{isensee2021nnu} & 33.10& - & - & - & - & -  \\ 
SwinUNETR~\citep{hatamizadeh2022swin} & 43.45  & \blue{\textbf{62.60}} & \blue{\textbf{90.32}} & {62.73} & \blue{\textbf{9.68}} & 37.27 \\
\hline 
\multicolumn{1}{c}{} & \multicolumn{6}{c}{SwinUNETR with different pre-train} 
\\ \hline
With pre-train in~\citep{tang2022self} & 44.06  & 57.79 & \lightblue{\textbf{89.25}} & 73.64 & 
\lightblue{\textbf{10.75}} & 26.36
\\
 \cellcolor{mygray} With \textbf{ours}   &
\cellcolor{mygray}\blue{\textbf{48.20}} & \cellcolor{mygray}\lightblue{\textbf{61.16}} & \cellcolor{mygray}{{88.17}} & \cellcolor{mygray}\blue{\textbf{77.27}} & \cellcolor{mygray}{11.83} & \cellcolor{mygray}\blue{\textbf{22.72}}
\\ \hline
\end{tabular}%
\caption{\textbf{Segmentation results of PET and CT} on AutoPET-II: 
Comparison between the previous method and ours.
The best results are highlighted in \blue{blue}. 
}
\vspace{-0.5cm}
\label{tab:seg_AutoPET-II}
\end{table*}

\subsection{Modality transfer results on BRATS22.}
\begin{figure*}[h]
\begin{center}
\includegraphics[width=\linewidth]{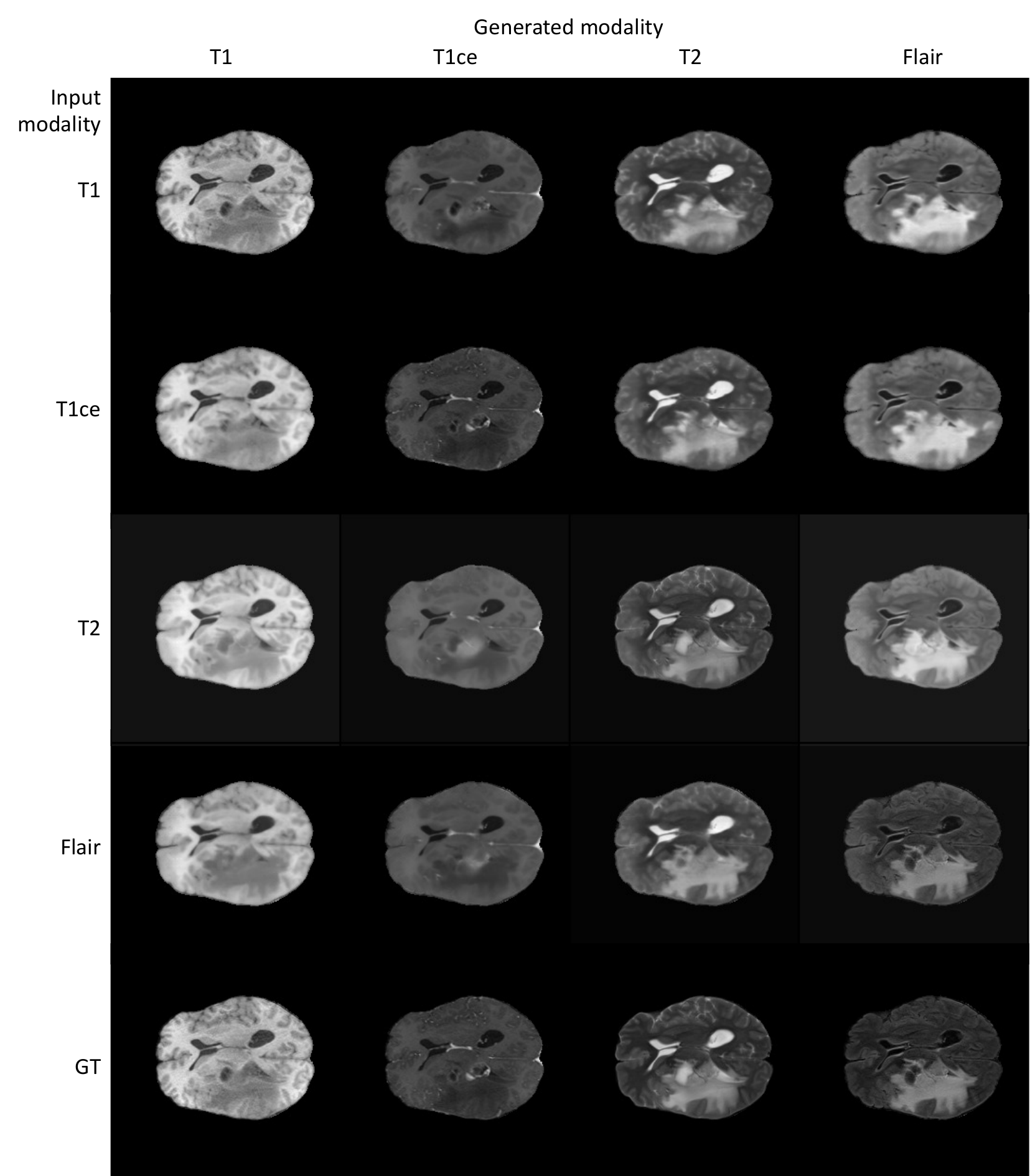}
\end{center}
\caption{Generated images of our proposed method: slices across ventricles.}
\label{fig:app_MRI_gen}
\end{figure*}

\begin{figure*}[h]
\begin{center}
\includegraphics[width=\linewidth]{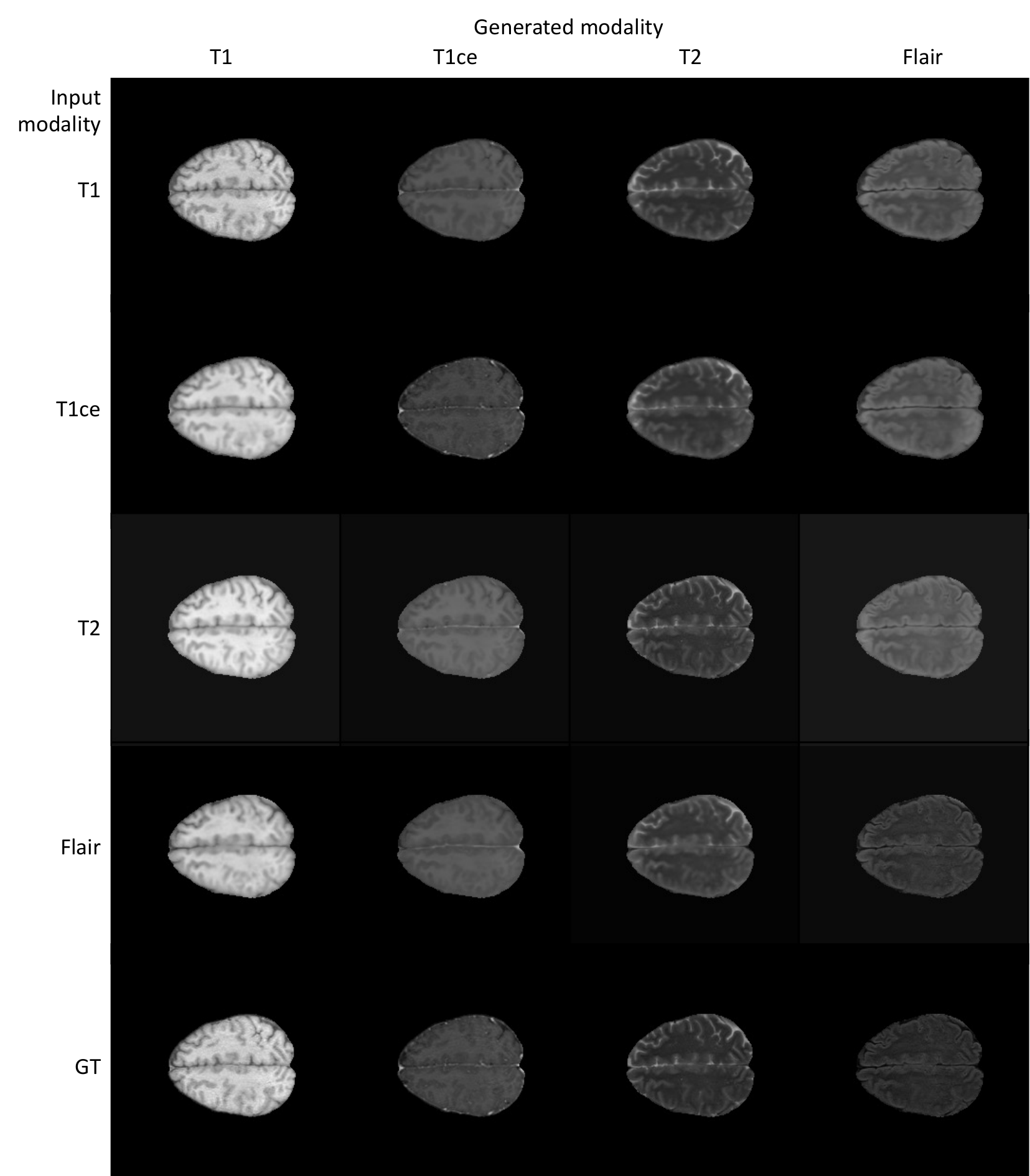}
\end{center}
\caption{Generated images of our proposed method: slices across cerebral sulcus.}
\label{fig:app_MRI_gen1}
\end{figure*}

\begin{figure*}[h]
\begin{center}
\includegraphics[width=\linewidth]{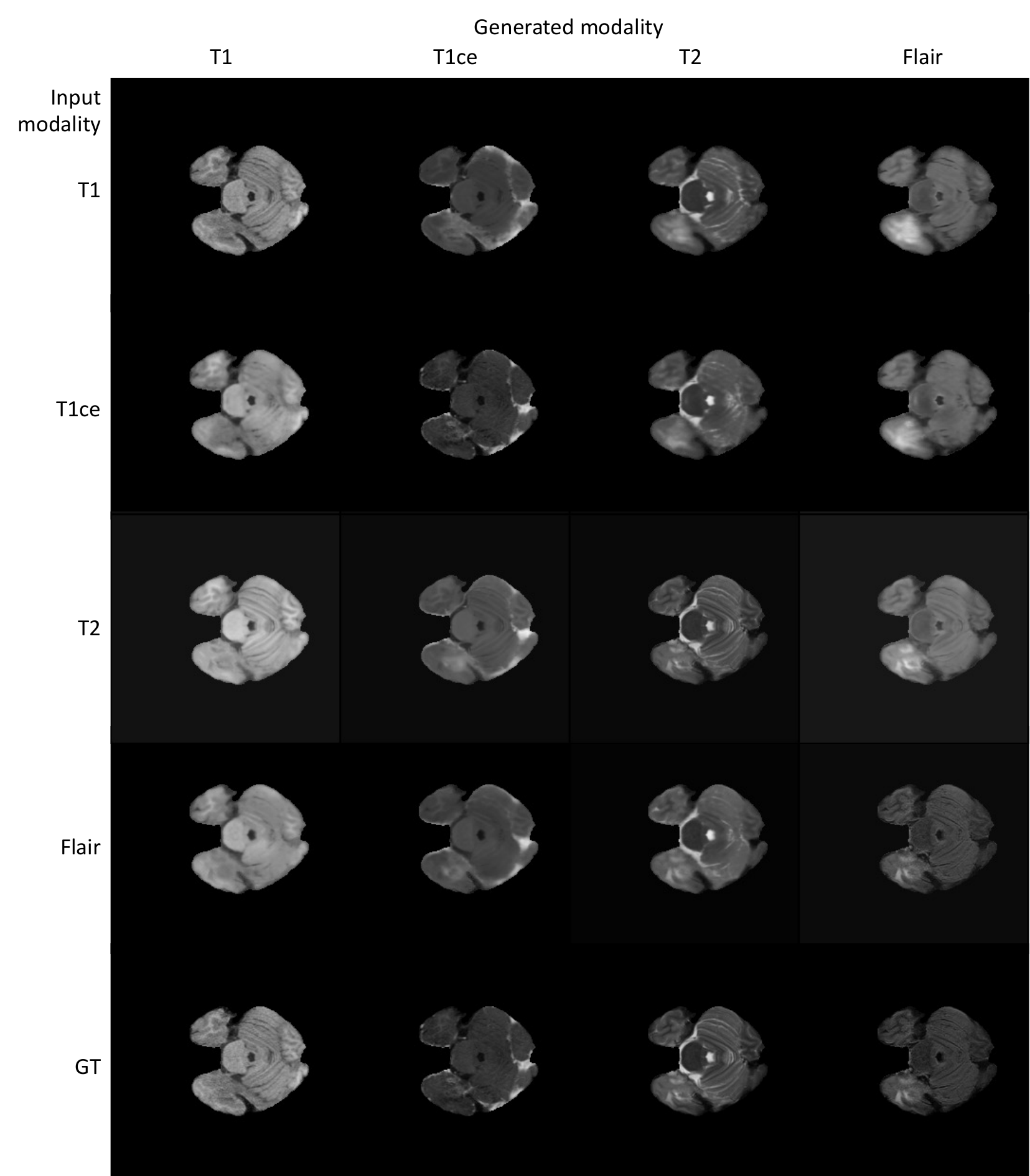}
\end{center}
\caption{Generated images of our proposed method: slices across the cerebellar hemisphere. Our method can generate defined cerebellar folia.}
\label{fig:app_MRI_gen2}
\end{figure*}

\cref{tab:app_gen_BRATS23_highlight} and \cref{tab:app_gen_BRATS23_main} presents the generation result with standard derivations. The results of our method and SwinUNETR are produced by ourselves, while the rest of the results are gathered from \cite{kim2024adaptive}. 
Generated examples are presented in \cref{fig:app_MRI_gen,fig:app_MRI_gen1,fig:app_MRI_gen2}.

\begin{table*}[t]

\resizebox{\textwidth}{!}{%
\begin{tabular}{cc|lll|lll}
\hline
 & \multicolumn{1}{c|}{\textbf{Task}} & \multicolumn{3}{c|}{\textbf{T1→T2}} & \multicolumn{3}{c}{\textbf{T2 → Flair}} \\ \hline
\multicolumn{1}{l}{Dimension} & Method & \multicolumn{1}{c}{PSNR↑} & \multicolumn{1}{c}{NMSE↓}  & \multicolumn{1}{c|}{SSIM↑} & \multicolumn{1}{c}{PSNR↑} & \multicolumn{1}{c}{NMSE↓} & \multicolumn{1}{c}{SSIM↑} \\ \hline
 & Pix2Pix & 24.624 ± 0.962 & 0.109 ± 0.028 & 0.874 ± 0.015 & 24.361 ± 1.061 & 0.117 ± 0.021 & 0.846 ± 0.019 \\
 & CycleGAN & 23.535 ± 1.334 & 0.155 ± 0.035 & 0.837 ± 0.028 & 23.418 ± 0.944 & 0.164 ± 0.033 & 0.825 ± 0.035 \\
 & NICEGAN & 23.721 ± 1.136 & 0.148 ± 0.029 & 0.840 ± 0.029 & 23.643 ± 1.045 & 0.148 ± 0.022 & 0.829 ± 0.033 \\
 & RegGAN & 24.884 ± 0.991 & 0.094 ± 0.024 & 0.881 ± 0.017 & 24.576 ± 1.073 & 0.112 ± 0.022 & 0.852 ± 0.028 \\
\multicolumn{1}{l}{\multirow{-5}{*}{2D}} & ResViT & 25.578 ± 0.812 & 0.088 ± 0.021 & 0.895 ± 0.018 & 24.825 ± 1.030 & 0.108 ± 0.018 & 0.861 ± 0.021 \\ \hline
 & CycleGAN & 25.181 ± 0.861 & 0.097 ± 0.031 & 0.887 ± 0.012 & 24.602 ± 1.181 & 0.113 ± 0.021 & 0.854 ± 0.018 \\
 & Pix2Pix & 23.740 ± 1.198 & 0.138 ± 0.032 & 0.835 ± 0.019 & 23.508 ± 1.301 & 0.152 ± 0.039 & 0.822 ± 0.024 \\
 & EaGAN & 24.884 ± 0.991 & 0.094 ± 0.024 & 0.881 ± 0.017 & 24.576 ± 1.073 & 0.112 ± 0.022 & 0.852 ± 0.028 \\
 & MS-SPADE & 25.818 ± 0.857 & 0.079 ± 0.016 & 0.904 ± 0.012 & 25.074 ± 1.085 & 0.098 ± 0.021 & 0.867 ± 0.018 \\
\multicolumn{1}{l}{\multirow{-5}{*}{3D}} & \cellcolor{mygray}\textbf{Ours} & \cellcolor{mygray}\blue{\textbf{30.756}} ± 1.950 & \cellcolor{mygray}\blue{\textbf{0.065}} ± 0.034 & 
\cellcolor{mygray}\blue{\textbf{0.944}} ± 0.031 & \cellcolor{mygray}\blue{\textbf{32.224}} ± 2.518 & \cellcolor{mygray}\blue{\textbf{0.046}} ± 0.029 & \cellcolor{mygray}\blue{\textbf{0.941}} ± 0.025 \\ \hline
\end{tabular}%
}
\caption{\textbf{Modality transfer results of MRI} on BRATS23: Comparison between previous methods and our method. 
The best results are highlighted in \blue{blue}.}
\label{tab:app_gen_BRATS23_highlight}
\end{table*}

\begin{table*}[t]

\resizebox{\textwidth}{!}{%
\begin{tabular}{cc|ccc|ccc|ccc|ccc}
\hline
\multicolumn{1}{l}{} & Target & \multicolumn{3}{c|}{T1} & \multicolumn{3}{c|}{T1ce} & \multicolumn{3}{c|}{T2} & \multicolumn{3}{c}{Flair} \\ \cline{3-14} 
\multicolumn{1}{l}{\textbf{Source}} & \textbf{} & PSNR↑ & NMSE↓ & SSIM↑ & PSNR↑ & NMSE↓ & SSIM↑ & PSNR↑ & NMSE↓ & SSIM↑ & PSNR↑ & NMSE↓ & SSIM↑ \\ \hline
\multicolumn{1}{c|}{} & \textbf{SwinUNETR} & \textbf{32.815} & \textbf{0.092} & \textbf{0.941} & \textbf{31.655} & \textbf{0.202} & \textbf{0.912} & \textbf{24.650} & \textbf{0.361} & \textbf{0.857} & \textbf{27.593} & \textbf{0.202} & \textbf{0.883} \\
\multicolumn{1}{c|}{} & {Std.} & \textit{0.968} & \textit{0.043} & \textit{0.049} & \textit{1.062} & \textit{0.067} & \textit{0.052} & \textit{1.008} & \textit{0.069} & \textit{0.077} & \textit{1.144} & \textit{0.072} & \textit{0.050} \\
\multicolumn{1}{c|}{} & \textbf{MS-SPADE} & \textbf{29.001} & \textbf{0.055} & \textbf{0.942} & \textbf{26.119} & \textbf{0.078} & \textbf{0.912} & \textbf{25.818} & \textbf{0.103} & \textbf{0.904} & \textbf{24.842} & \textbf{0.113} & \textbf{0.859} \\
\multicolumn{1}{c|}{} & {Std.} & \textit{0.643} & \textit{0.025} & \textit{0.022} & \textit{0.816} & \textit{0.022} & \textit{0.015} & \textit{0.857} & \textit{0.030} & \textit{0.014} & \textit{0.728} & \textit{0.034} & \textit{0.019} \\
\multicolumn{1}{c|}{} &  \cellcolor{mygray}\textbf{Ours} &  \cellcolor{mygray}\textbf{43.472} &  \cellcolor{mygray}\textbf{0.003} &  \cellcolor{mygray}\textbf{0.996} &  \cellcolor{mygray}\textbf{34.547} &  \cellcolor{mygray}\textbf{0.045} &  \cellcolor{mygray}\textbf{0.955} &  \cellcolor{mygray}\textbf{30.756} &  \cellcolor{mygray}\textbf{0.065} &  \cellcolor{mygray}\textbf{0.944} &  \cellcolor{mygray}\textbf{31.693} &  \cellcolor{mygray}\textbf{0.049} &  \cellcolor{mygray}\textbf{0.937} \\
\multicolumn{1}{c|}{\multirow{-6}{*}{\textbf{T1}}} &  \cellcolor{mygray}{Std.} &  \cellcolor{mygray}\textit{2.495} &  \cellcolor{mygray}\textit{0.004} &  \cellcolor{mygray}\textit{0.011} &  \cellcolor{mygray}\textit{1.956} &  \cellcolor{mygray}\textit{0.030} &  \cellcolor{mygray}\textit{0.018} &  \cellcolor{mygray}\textit{1.950} &  \cellcolor{mygray}\textit{0.034} &  \cellcolor{mygray}\textit{0.031} &  \cellcolor{mygray}\textit{2.287} &  \cellcolor{mygray}\textit{0.024} &  \cellcolor{mygray}\textit{0.019} \\ \hline
\multicolumn{1}{c|}{} & \textbf{SwinUNETR} & \textbf{32.456} & \textbf{0.100} & \textbf{0.929} & \textbf{33.001} & \textbf{0.156} & \textbf{0.926} & \textbf{25.125} & \textbf{0.366} & \textbf{0.859} & \textbf{27.699} & \textbf{0.211} & \textbf{0.882} \\
\multicolumn{1}{c|}{} & {Std.} & \textit{1.018} & \textit{0.044} & \textit{0.048} & \textit{0.889} & \textit{0.055} & \textit{0.051} & \textit{0.964} & \textit{0.071} & \textit{0.074} & \textit{1.129} & \textit{0.071} & \textit{0.049} \\
\multicolumn{1}{c|}{} & \textbf{MS-SPADE} & \textbf{26.228} & \textbf{0.076} & \textbf{0.922} & \textbf{28.759} & \textbf{0.060} & \textbf{0.937} & \textbf{25.990} & \textbf{0.092} & \textbf{0.907} & \textbf{25.204} & \textbf{0.092} & \textbf{0.881} \\
\multicolumn{1}{c|}{} & {Std.} & \textit{0.794} & \textit{0.027} & \textit{0.033} & \textit{0.885} & \textit{0.019} & \textit{0.015} & \textit{0.859} & \textit{0.032} & \textit{0.908} & \textit{0.811} & \textit{0.050} & \textit{0.037} \\
\multicolumn{1}{c|}{} &  \cellcolor{mygray}\textbf{Ours} &  \cellcolor{mygray}\textbf{34.077} &  \cellcolor{mygray}\textbf{0.020} &  \cellcolor{mygray}\textbf{0.962} &  \cellcolor{mygray}\textbf{46.663} &  \cellcolor{mygray}\textbf{0.003} &  \cellcolor{mygray}\textbf{0.996} &  \cellcolor{mygray}\textbf{30.775} &  \cellcolor{mygray}\textbf{0.063} &  \cellcolor{mygray}\textbf{0.942} &  \cellcolor{mygray}\textbf{32.224} &  \cellcolor{mygray}\textbf{0.046} &  \cellcolor{mygray}\textbf{0.941} \\
\multicolumn{1}{c|}{\multirow{-6}{*}{\textbf{T1ce}}} &  \cellcolor{mygray}{Std.} &  \cellcolor{mygray}\textit{2.484} &  \cellcolor{mygray}\textit{0.012} &  \cellcolor{mygray}\textit{0.017} &  \cellcolor{mygray}\textit{3.240} &  \cellcolor{mygray}\textit{0.004} &  \cellcolor{mygray}\textit{0.008} &  \cellcolor{mygray}\textit{1.812} &  \cellcolor{mygray}\textit{0.030} &  \cellcolor{mygray}\textit{0.028} &  \cellcolor{mygray}\textit{2.518} &  \cellcolor{mygray}\textit{0.029} &  \cellcolor{mygray}\textit{0.025} \\ \hline
\multicolumn{1}{c|}{} & \textbf{SwinUNETR} & \textbf{30.102} & \textbf{0.171} & \textbf{0.896} & \textbf{30.354} & \textbf{0.283} & \textbf{0.883} & \textbf{26.831} & \textbf{0.268} & \textbf{0.887} & \textbf{27.234} & \textbf{0.242} & \textbf{0.872} \\
\multicolumn{1}{c|}{} & {Std.} & \textit{1.405} & \textit{0.056} & \textit{0.050} & \textit{1.249} & \textit{0.086} & \textit{0.054} & \textit{1.144} & \textit{0.054} & \textit{0.075} & \textit{1.154} & \textit{0.073} & \textit{0.051} \\
\multicolumn{1}{c|}{} & \textbf{MS-SPADE} & \textbf{25.422} & \textbf{0.085} & \textbf{0.908} & \textbf{25.234} & \textbf{0.087} & \textbf{0.895} & \textbf{29.230} & \textbf{0.048} & \textbf{0.942} & \textbf{25.074} & \textbf{0.098} & \textbf{0.867} \\
\multicolumn{1}{c|}{} & {Std.} & \textit{0.852} & \textit{0.026} & \textit{0.020} & \textit{1.152} & \textit{0.034} & \textit{0.025} & \textit{0.720} & \textit{0.018} & \textit{0.915} & \textit{1.085} & \textit{0.021} & \textit{0.018} \\
\multicolumn{1}{c|}{} &  \cellcolor{mygray}\textbf{Ours} &  \cellcolor{mygray}\textbf{32.646} &  \cellcolor{mygray}\textbf{0.028} &  \cellcolor{mygray}\textbf{0.955} &  \cellcolor{mygray}\textbf{33.857} &  \cellcolor{mygray}\textbf{0.051} &  \cellcolor{mygray}\textbf{0.949} &  \cellcolor{mygray}\textbf{43.653} &  \cellcolor{mygray}\textbf{0.006} &  \cellcolor{mygray}\textbf{0.991} &  \cellcolor{mygray}\textbf{32.224} &  \cellcolor{mygray}\textbf{0.046} &  \cellcolor{mygray}\textbf{0.941} \\
\multicolumn{1}{c|}{\multirow{-6}{*}{\textbf{T2}}} &  \cellcolor{mygray}{Std.} &  \cellcolor{mygray}\textit{2.391} &  \cellcolor{mygray}\textit{0.028} &  \cellcolor{mygray}\textit{0.028} &  \cellcolor{mygray}\textit{1.925} &  \cellcolor{mygray}\textit{0.040} &  \cellcolor{mygray}\textit{0.027} &  \cellcolor{mygray}\textit{3.467} &  \cellcolor{mygray}\textit{0.024} &  \cellcolor{mygray}\textit{0.038} &  \cellcolor{mygray}\textit{2.518} &  \cellcolor{mygray}\textit{0.029} &  \cellcolor{mygray}\textit{0.025} \\ \hline
\multicolumn{1}{c|}{} & \textbf{SwinUNETR} & \textbf{31.371} & \textbf{0.135} & \textbf{0.916} & \textbf{31.285} & \textbf{0.240} & \textbf{0.905} & \textbf{25.579} & \textbf{0.338} & \textbf{0.867} & \textbf{29.092} & \textbf{0.148} & \textbf{0.923} \\
\multicolumn{1}{c|}{} & {Std.} & \textit{1.198} & \textit{0.051} & \textit{0.054} & \textit{1.161} & \textit{0.077} & \textit{0.053} & \textit{0.956} & \textit{0.064} & \textit{0.073} & \textit{0.974} & \textit{0.055} & \textit{0.049} \\
\multicolumn{1}{c|}{} & \textbf{MS-SPADE} & \textbf{25.186} & \textbf{0.090} & \textbf{0.905} & \textbf{25.899} & \textbf{0.094} & \textbf{0.906} & \textbf{26.146} & \textbf{0.086} & \textbf{0.913} & \textbf{28.608} & \textbf{0.058} & \textbf{0.938} \\
\multicolumn{1}{c|}{} & {Std.} & \textit{0.759} & \textit{0.028} & \textit{0.048} & \textit{1.039} & \textit{0.025} & \textit{0.027} & \textit{0.636} & \textit{0.028} & \textit{0.944} & \textit{0.769} & \textit{0.025} & \textit{0.028} \\
\multicolumn{1}{c|}{} &  \cellcolor{mygray}\textbf{Ours} &  \cellcolor{mygray}\textbf{32.752} &  \cellcolor{mygray}\textbf{0.026} &  \cellcolor{mygray}\textbf{0.951} &  \cellcolor{mygray}\textbf{33.471} &  \cellcolor{mygray}\textbf{0.055} &  \cellcolor{mygray}\textbf{0.944} &  \cellcolor{mygray}\textbf{30.571} &  \cellcolor{mygray}\textbf{0.068} &  \cellcolor{mygray}\textbf{0.940} &  \cellcolor{mygray}\textbf{43.624} &  \cellcolor{mygray}\textbf{0.004} &  \cellcolor{mygray}\textbf{0.995} \\
\multicolumn{1}{c|}{\multirow{-6}{*}{\textbf{Flair}}} &  \cellcolor{mygray}{Std.} &  \cellcolor{mygray}\textit{2.399} &  \cellcolor{mygray}\textit{0.020} &  \cellcolor{mygray}\textit{0.022} &  \cellcolor{mygray}\textit{1.634} &  \cellcolor{mygray}\textit{0.035} &  \cellcolor{mygray}\textit{0.021} &  \cellcolor{mygray}\textit{1.951} &  \cellcolor{mygray}\textit{0.035} &  \cellcolor{mygray}\textit{0.034} &  \cellcolor{mygray}\textit{2.441} &  \cellcolor{mygray}\textit{0.008} &  \cellcolor{mygray}\textit{0.013} \\
\hline
\end{tabular}%
}
\caption{\textbf{Modality transfer results of MRI} on BRATS23: The averaged results with standard derivations of metrics between all modalities. 
}
\label{tab:app_gen_BRATS23_main}
\end{table*}

\subsection{Missing modality segmentation results on BRATS18.}
We provide detailed segmentation results on BRATS18 as \cref{tab:app_missing_seg_brats18_main}.

\subsection{Comparison with non-personalized methods.}
\begin{figure}
    \centering
      \captionof{figure}{TSNE of latent features of PET and CT images obtained under different applied constraints. Downstream performances are noted.}
      \label{fig:re_vis}
      \vspace{-0.3cm}
    \includegraphics[width=0.9\linewidth]{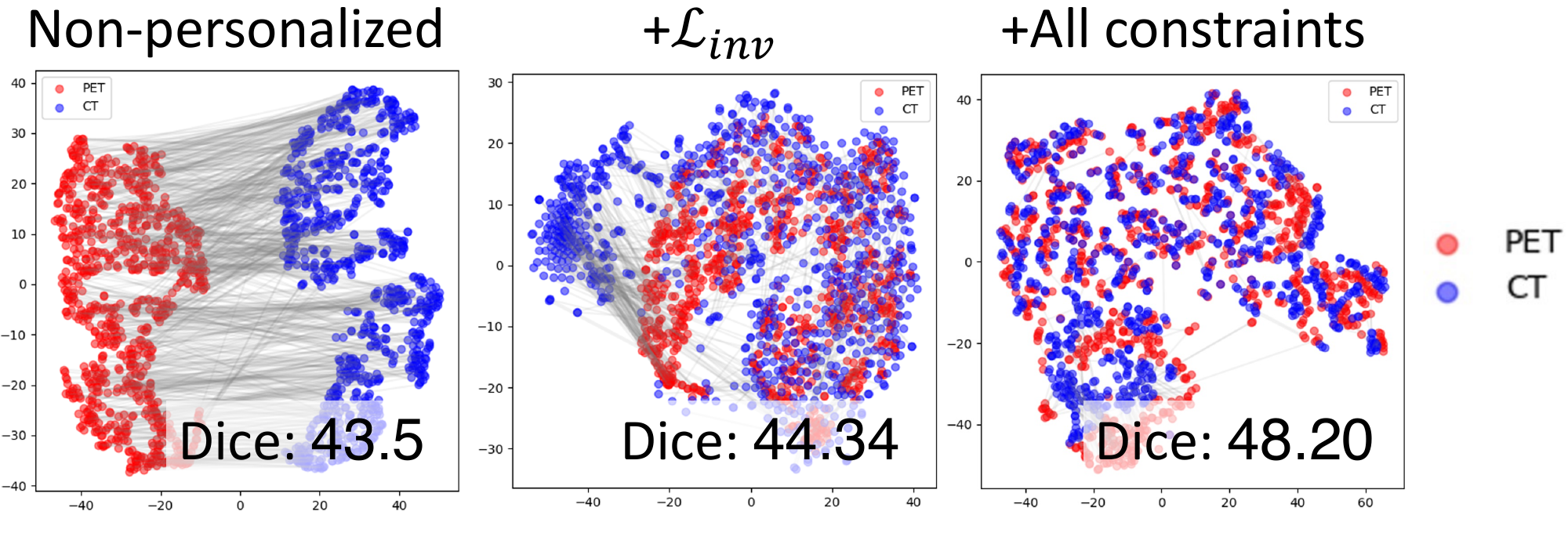}
\end{figure}

We provide further visual evidence in \cref{fig:re_vis} for an in-depth analysis.  \cref{fig:re_vis} shows that applying our constraints aligns CT and PET representations more closely, indicating smaller $ d_{\mathcal{G}\Delta\mathcal{G}}(\mathcal{M} | \mathcal{H}^\mathcal{S}, \mathcal{M} | \mathcal{H}^\mathcal{U})$ in Paper Eq.~(10), yielding tightened bounds and better downstream performance (Dice:48.20), in comparison to the non-personalized baseline (Dice:43.5). This further supports our theoretical analysis. 

\begin{table*}[ht]

\resizebox{\textwidth}{!}{%
\begin{tabular}{c|c|cccc|cccccc|cccc|c}
\hline
\multicolumn{2}{c|}{Missing Num} & \multicolumn{4}{c|}{=3} & \multicolumn{6}{c|}{=2} & \multicolumn{4}{c|}{=1} & =0 \\ 
\hline
 & flair & \multicolumn{1}{l}{\cellcolor[HTML]{AEAAAA}} & \multicolumn{1}{l}{} & \multicolumn{1}{l}{} & \multicolumn{1}{l|}{} & \multicolumn{1}{l}{\cellcolor[HTML]{AEAAAA}} & \multicolumn{1}{l}{\cellcolor[HTML]{AEAAAA}} & \multicolumn{1}{l}{} & \multicolumn{1}{l}{} & \multicolumn{1}{l}{\cellcolor[HTML]{AEAAAA}} & \multicolumn{1}{l|}{} & \multicolumn{1}{l}{\cellcolor[HTML]{AEAAAA}} & \multicolumn{1}{l}{\cellcolor[HTML]{AEAAAA}} & \multicolumn{1}{l}{\cellcolor[HTML]{AEAAAA}} & \multicolumn{1}{l|}{} & \multicolumn{1}{l}{\cellcolor[HTML]{AEAAAA}} \\
 & T1 & \multicolumn{1}{l}{} & \multicolumn{1}{l}{\cellcolor[HTML]{AEAAAA}} & \multicolumn{1}{l}{} & \multicolumn{1}{l|}{} & \multicolumn{1}{l}{\cellcolor[HTML]{AEAAAA}} & \multicolumn{1}{l}{} & \multicolumn{1}{l}{\cellcolor[HTML]{AEAAAA}} & \multicolumn{1}{l}{\cellcolor[HTML]{AEAAAA}} & \multicolumn{1}{l}{} & \multicolumn{1}{l|}{} & \multicolumn{1}{l}{\cellcolor[HTML]{AEAAAA}} & \multicolumn{1}{l}{\cellcolor[HTML]{AEAAAA}} & \multicolumn{1}{l}{} & \multicolumn{1}{l|}{\cellcolor[HTML]{AEAAAA}} & \multicolumn{1}{l}{\cellcolor[HTML]{AEAAAA}} \\
 & T1ce & \multicolumn{1}{l}{} & \multicolumn{1}{l}{} & \multicolumn{1}{l}{\cellcolor[HTML]{AEAAAA}} & \multicolumn{1}{l|}{} & \multicolumn{1}{l}{} & \multicolumn{1}{l}{\cellcolor[HTML]{AEAAAA}} & \multicolumn{1}{l}{} & \multicolumn{1}{l}{\cellcolor[HTML]{AEAAAA}} & \multicolumn{1}{l}{} & \multicolumn{1}{l|}{\cellcolor[HTML]{AEAAAA}} & \multicolumn{1}{l}{\cellcolor[HTML]{AEAAAA}} & \multicolumn{1}{l}{} & \multicolumn{1}{l}{\cellcolor[HTML]{AEAAAA}} & \multicolumn{1}{l|}{\cellcolor[HTML]{AEAAAA}} & \multicolumn{1}{l}{\cellcolor[HTML]{AEAAAA}} \\
\multirow{-4}{*}{Modality} & T2 & \multicolumn{1}{l}{} & \multicolumn{1}{l}{} & \multicolumn{1}{l}{} & \multicolumn{1}{l|}{\cellcolor[HTML]{AEAAAA}} & \multicolumn{1}{l}{} & \multicolumn{1}{l}{} & \multicolumn{1}{l}{\cellcolor[HTML]{AEAAAA}} & \multicolumn{1}{l}{} & \multicolumn{1}{l}{\cellcolor[HTML]{AEAAAA}} & \multicolumn{1}{l|}{\cellcolor[HTML]{AEAAAA}} & \multicolumn{1}{l}{} & \multicolumn{1}{l}{\cellcolor[HTML]{AEAAAA}} & \multicolumn{1}{l}{\cellcolor[HTML]{AEAAAA}} & \multicolumn{1}{l|}{\cellcolor[HTML]{AEAAAA}} & \multicolumn{1}{l}{\cellcolor[HTML]{AEAAAA}} \\ \hline
 & SPA & 65.86 & 65.27 & 78.26 & 66.4 & 72.99 & 83.23 & 70.66 & 81.25 & 70.66 & 80.63 & 83.22 & 73.89 & 83.36 & 82.05 & 83.4 \\
 & M3AE & 69.4 & 65.45 & 79.12 & 71.84 & 79.9 & 70.45 & 82.79 & 81.17 & 71.62 & 73.35 & 81.78 & 82.42 & 73.31 & 81.61 & 82.22 \\
 & mmFormer & 67.8 & 77.32 & 64.56 & 64.08 & 81.51 & 79.43 & 69.14 & 70.63 & 68.6 & 80.75 & 81.75 & 70.92 & 81.74 & 81.55 & 82.23 \\
 & RFNET & 64.03 & 74.53 & 58.63 & 61.95 & 79.2 & 77.45 & 69.25 & 67.48 & 67.98 & 78.85 & 80.15 & 70.75 & 79.4 & 80.15 & 80.29 \\
 & M2F & 65.79 & 63.29 & 77.31 & 63.64 & 70.38 & 79.93 & 68.01 & 79.62 & 67.68 & 79.37 & 80.65 & 69.73 & 80.01 & 79.53 & 80.34 \\
\multirow{-6}{*}{Tumour Core} & \cellcolor{mygray}\textbf{Ours} & \cellcolor{mygray}75.83 & \cellcolor{mygray}71.2 & \cellcolor{mygray}75.29 & \cellcolor{mygray}75.71 & \cellcolor{mygray}80.66 & \cellcolor{mygray}83.6 & \cellcolor{mygray}79.23 & \cellcolor{mygray}74.83 & \cellcolor{mygray}79.51 & \cellcolor{mygray}79.52 & \cellcolor{mygray}83.92 & \cellcolor{mygray}82.78 & \cellcolor{mygray}86.65 & \cellcolor{mygray}81.22 & \cellcolor{mygray}86.72 \\ \hline
 & SPA & 39.85 & 41.39 & 70.43 & 41.72 & 45.99 & 73.07 & 45.25 & 72.87 & 45.25 & 72.59 & 73.52 & 47.56 & 73.01 & 73.55 & 73.65 \\
 & M3AE & 37 & 38.41 & 75.8 & 44.22 & 78.09 & 45.2 & 79.36 & 78.16 & 41.71 & 48.12 & 79.14 & 80.06 & 47.63 & 79.31 & 79.91 \\
 & mmFormer & 40.08 & 72.19 & 38.89 & 37.23 & 73.11 & 73.06 & 40.64 & 42.27 & 43.65 & 75.56 & 43.34 & 81.74 & 73.36 & 75.31 & 73.4 \\
 & RFNET & 38.69 & 69.22 & 30.89 & 33.56 & 71.4 & 70.9 & 38.53 & 41.91 & 40.9 & 69.51 & 71.61 & 43.37 & 71.17 & 74.2 & 73.79 \\
 & M2F & 37.99 & 37.79 & 71.74 & 39.28 & 43.37 & 74.66 & 45.42 & 73.48 & 43.5 & 73.48 & 73.56 & 45.93 & 73.15 & 74.03 & 75.26 \\
\multirow{-6}{*}{Enhancing tumour} & \cellcolor{mygray}\textbf{Ours} & \cellcolor{mygray}67.45 & \cellcolor{mygray}54.83 & \cellcolor{mygray}70.86 & \cellcolor{mygray}47.63 & \cellcolor{mygray}69.38 & \cellcolor{mygray}52.91 & \cellcolor{mygray}70.1 & \cellcolor{mygray}59.45 & \cellcolor{mygray}67.44 & \cellcolor{mygray}63.91 & \cellcolor{mygray}70.79 & \cellcolor{mygray}57.78 & \cellcolor{mygray}69.42 & \cellcolor{mygray}59.76 & \cellcolor{mygray}70.64 \\ \hline
 & SPA & 85.77 & 72.69 & 71.95 & 80.4 & 87.82 & 87.97 & 88.27 & 75.57 & 88.27 & 81.8 & 88.3 & 88.78 & 89.06 & 82.87 & 89.27 \\
 & M3AE & 87.78 & 74.69 & 74.91 & 84.43 & 76.09 & 84.48 & 89.63 & 84.4 & 88.64 & 88.91 & 84.04 & 89.29 & 88.58 & 88.45 & 88.26 \\
 & mmFormer & 84.09 & 72.85 & 73.37 & 85.6 & 85.97 & 76.93 & 87.09 & 86.09 & 87.55 & 87.94 & 88.36 & 88.16 & 88.74 & 85.96 & 89.03 \\
 & RFNET & 80.52 & 67.06 & 68.42 & 82.96 & 82.57 & 71.97 & 85.82 & 83.25 & 86 & 84.94 & 86.06 & 86.53 & 86.34 & 83.61 & 86.82 \\
 & M2F & 85.72 & 72.48 & 71.78 & 82.53 & 87.73 & 87.66 & 84.35 & 76.03 & 87.69 & 84.27 & 88.17 & 88.22 & 88.47 & 84.32 & 88.72 \\
\multirow{-6}{*}{Whole tumour} & \cellcolor{mygray}\textbf{Ours} & \cellcolor{mygray}89.23 & \cellcolor{mygray}81.73 & \cellcolor{mygray}82.26 & \cellcolor{mygray}87.45 & \cellcolor{mygray}89.74 & \cellcolor{mygray}89.03 & \cellcolor{mygray}88.00 & \cellcolor{mygray}81.92 & \cellcolor{mygray}89.72 & \cellcolor{mygray}86.27 & \cellcolor{mygray}89.12 & \cellcolor{mygray}90.5 & \cellcolor{mygray}91.25 & \cellcolor{mygray}87.1 & \cellcolor{mygray}91.19 \\ \hline
\end{tabular}%
}
\caption{\textbf{Missing modality segmentation results of MRI} on BRATS18: Num denotes the number of missing modalities for different settings. The used modalities are highlighted with gray boxes and the missing ones remain as blank. 
The results of each setting are presented accordingly.
}
\label{tab:app_missing_seg_brats18_main}
\end{table*}


\end{document}